\pdfoutput=1
\documentclass[11pt]{article}

\usepackage{EMNLP2022}

\usepackage{times}
\usepackage{latexsym}
\usepackage{amsfonts} 
\usepackage{amsmath}
\usepackage{graphicx}
\usepackage{comment}
\usepackage{amssymb}

\usepackage[T1]{fontenc}
\usepackage[utf8]{inputenc}
\usepackage{booktabs}

\usepackage{microtype}
\usepackage{blindtext}
\usepackage{subfiles}

 \newif\ifcomment\commentfalse
\usepackage[a-1b]{pdfx}

\usepackage{framed}
\usepackage{mdwlist}
\usepackage{siunitx}
\usepackage{latexsym}
\usepackage{colortbl}
\usepackage{xcolor}
\usepackage{nicefrac}
\usepackage{booktabs}
\usepackage{fnpct}
\usepackage{amsfonts}
\usepackage[T1]{fontenc}
\usepackage{bold-extra}
\usepackage{amsmath}
\usepackage{amssymb}
\usepackage{bm}
\usepackage{graphicx}
\usepackage{mathtools}
\usepackage{microtype}
\usepackage{multirow}
\usepackage{multicol}
\usepackage{xpatch}
\usepackage{latexsym,comment}
\usepackage[normalem]{ulem}

\newcommand*{\missingreference}{{\Huge \colorbox{red}{?reference?}}}
\newcommand*{\missingcitation}{{\Huge \colorbox{red}{?citation?}}}

\makeatletter
\xpatchcmd{\@setref}{\bfseries}{\missingreference}{}{}
\def\@citex[#1]#2{\leavevmode
    \let\@citea\@empty
    \@cite{\@for\@citeb:=#2\do
        {\@citea\def\@citea{,\penalty\@m\ }%
            \edef\@citeb{\expandafter\@firstofone\@citeb\@empty}%
            \if@filesw\immediate\write\@auxout{\string\citation{\@citeb}}\fi
            \@ifundefined{b@\@citeb}{\hbox{\reset@font\missingcitation}%
                \G@refundefinedtrue
                \@latex@warning
                {Citation `\@citeb' on page \thepage \space undefined}}%
            {\@cite@ofmt{\csname b@\@citeb\endcsname}}}}{#1}}
\makeatother

\newcommand{\gem}[1]{\mbox{\textsc{gem}}}
\newcommand{\abr}[1]{\textsc{#1}}

\newcommand{\g}{\, | \,}

\newcommand{\hidetext}[1]{}
\newcommand{\ignore}[1]{}

\ifcomment
    \newcommand{\pinaforecomment}[3]{\colorbox{#1}{\parbox{.8\linewidth}{#2: #3}}}

    \newcommand{\prtodo}[1]{\pinaforecomment{lightblue}{pr}{#1}}
    \newcommand{\prtodoi}[1]{\pinaforecomment{lightblue}{pr}{#1}}
\else
    \newcommand{\pinaforecomment}[3]{}
    \newcommand{\prtodo}[1]{}
    \newcommand{\prtodoi}[1]{}
\fi

\newcommand{\jbgcomment}[1]{\pinaforecomment{red}{JBG}{#1}}

\newcommand{\smallurl}[1]{ \begin{tiny}\url{#1}\end{tiny}}

\definecolor{lightblue}{HTML}{3cc7ea}
\definecolor{CUgold}{HTML}{CFB87C}
\definecolor{grey}{rgb}{0.95,0.95,0.95}
\definecolor{ceil}{rgb}{0.57, 0.63, 0.81}
\definecolor{UMDred}{HTML}{ed1c24}
\definecolor{UMDyellow}{HTML}{ffc20e}

\newcommand{\chenglei}[1]{
    \textcolor{magenta}{[Chenglei: #1]}
}

\newcommand{\sewon}[1]{
    \textcolor{teal}{[Sewon: #1]}
}

\newcommand{\metric}{\textsc{MacroCE}}
\newcommand{\name}{\textsc{ConsCal}}

\title{Re-Examining Calibration: The Case of Question Answering}

\author{
  Chenglei Si\\
  University of Maryland\\
  {\tt clsi@umd.edu} \\\And
  Chen Zhao\\
  New York University\\
  {\tt cz1285@nyu.edu} \\\AND
  Sewon Min\\
  University of Washington\\
  {\tt sewon@cs.washington.edu} \\\And 
  Jordan Boyd-Graber\\
  University of Maryland\\
  {\tt jbg@umiacs.umd.edu} \\\newline}

\begin{document}
\maketitle
\begin{abstract}
% \chenglei{Initial draft, feel free to edit.}
%\sewon{Important comment: abstract mentions nothing about ODQA?! \jbgcomment{added}}
%
For users to trust model predictions, they need to understand model outputs, particularly their
confidence---calibration aims to adjust (calibrate) models' confidence to
match expected accuracy.
We argue that the traditional calibration evaluation does not promote effective calibrations: for example, it can encourage always assigning a mediocre confidence score to all predictions,
which does not
help users distinguish correct predictions from wrong ones.
Building on those observations, we propose a new calibration metric, \metric{},
that better captures whether the model assigns low confidence to wrong
predictions and high confidence to correct predictions.
Focusing on the practical application of open-domain question answering, we
examine conventional calibration methods applied on the widely-used retriever-reader pipeline, all of which do not bring
significant gains under our new \metric{} metric.
Toward better calibration, we propose a new calibration method (\name{}) that
uses not just final model predictions but whether multiple model checkpoints make consistent predictions.
Altogether, we provide an alternative view of calibration along with a new metric, 
re-evaluation of existing calibration methods on our metric, and 
proposal of a more effective calibration method.~\footnote{Code available at: \url{https://github.com/NoviScl/calibrateQA}}

% \sewon{General comment about the abstract: I think the abstract should mention QA somewhere, given that the title contains QA and we mention QA early on in the introduction.}
% }
% \jbgcomment{In US English, we normally say ``toward'' rather than ``towards''.  I'm okay with UK English so long as we do it consistently}
%
% \chen{I add a sentence as a summary}
% \sewon{Feel like there should be a better name other than consistency calibration... } \chen{I add a placeholder, but we should have a new name}
%
% This new method, which we name as consistency calibration,  shows promise for better calibration. 
% While we mainly ground our discussion on the task of question answering in this paper, we expect our findings to extend beyond and be more widely applicable. 

% \chen{Do you all like this title?}
% \sewon{I think the title makes sense if we start the abstract/info with ODQA. Currently, it starts from inherent problem of calibration and then mention ODQA as an application (which I actually like much better!) and then now the title makes less sense.}
% \chenglei{Proposed a new title.}

\end{abstract}

\section{Introduction}
 %\chen{Chenglei: Could you run a style check script before Jordan making the last edit (e.g., before Thursday}
% structure (suggested by Chen): 
% 1. Pre-trained models are strong, but still make mistakes. Therefore we should decide when to abstain, based on the calibrated model predictions
% 2. This is specifically important in QA, as 1) it has real impacts in e.g., search engine; 2) It's more complex, contains multiple components 
% 3. metric is problematic
% 4.  and we propose an alternative 

% \sewon{TODO: use one of ODQA and \abr{odqa} consistently. Same for metrics, method name, etc.}
%\chen{Important: make all marcos consistent, we should talk about it on Thursday's meeting.}

%\chen{I prefer not citing papers from our group at the first paragraph of Intro}.

%While the predictions of large pre-trained language models (\plm) are ubiquitous~\cite{devlin-19,Brown2020LanguageMA}, 
While large pretrained language models have conquered many downstream
tasks~\cite{devlin-19,Brown2020LanguageMA}, it is sometimes unclear
when we should trust them since they often produce
false~\cite{TruthfulQA} or hallucinated~\cite{Maynez2020OnFA}
predictions.
% \jbgcomment{add some cites on hallucinations, etc.}
%
%% cz0612: I slightly reword it to make it more general
%This is an issue not just for users who need to decide whether to trust the outputs they get from digital assistants, but also for companies who reasonably want to censor low-confidence outputs from users or downstream systems.
This is important for both model deployment---where low-confidence
outputs can be censored---and end users who need to know whether to
trust a model output.
%
%
%models need to abstain from predictions that are not confident, since to get user trust, models should always avoid presenting prediction errors to users.
%Therefore, in addition to making predictions,
%
%users should decide when to abstain from model predictions. 
%
%these models are often viewed as a black-box, and the predictions are still
%far from perfect.
The solution is to make sure that models provide reliable confidence
estimates so that we can abstain from wrong predictions and trust the
right ones.
%, therefore users can decide when to abstain from model predictions. 
%to deploy these modeling progress into real scenarios, 
%
The prerequisite for such abstention is \textit{Model Calibration}: making
the confidence represent the actual likelihood of  being
correct~\cite{NiculescuMizil2005PredictingGP,Naeini2015ObtainingWC}.
Past work proposes post-hoc approaches to calibrate model onfidence
such as temperature scaling~\cite{guo2017calibration}, and can
effectively calibrate multi-class classification, evaluated by the
expected calibration error (\abr{ece}) metric.
%Past work has proposed post-hoc approaches to calibrate model predictions, such as temperature scaling~\cite{guo2017calibration}, which is designed for multi-class classification tasks.
%

% \begin{figure}[t]
%     \centering
%     \includegraphics[width=0.5\textwidth]{figures/Fig1.png}
%     \caption{Top: Temperature scaling puts correct and wrong predictions in the same confidence bucket (\textit{i.e.}, similar confidence). Bottom: We propose a new consistency calibration which assigns high confidence to correct predictions and low confidence to wrong predictions. Middle: Consistency calibration works by tracking whether the model makes the same prediction throughout the training trajectory. In this example we save three checkpoints in training. The checkpoints predict different answers for Q1 and so the final prediction is assigned a low confidence, in contrast the model predicts the same answer throughout training for Q2 and gets a high confidence.
%     \sewon{I'm not sure if this figure is good as a teaser figure. x-axis and y-axis are not very clear. It took me a while to understand the figure so I think the readers who haven't read the paper will take even longer. I feel like some combination of Figure 2 and 3, or a better illustration of consistency calibration method might be better.}
%     }
%     \jbgcomment{This figure would be more effective if it captured the idea of bins and the ``regression to the mean'' of ECE}
%     \label{fig:fig1}
% \end{figure}

 \begin{figure}[t]
 \centering
 \includegraphics[trim=0.6cm 0.5cm 1.0cm 0.5cm,clip=true,width=1.01\linewidth]{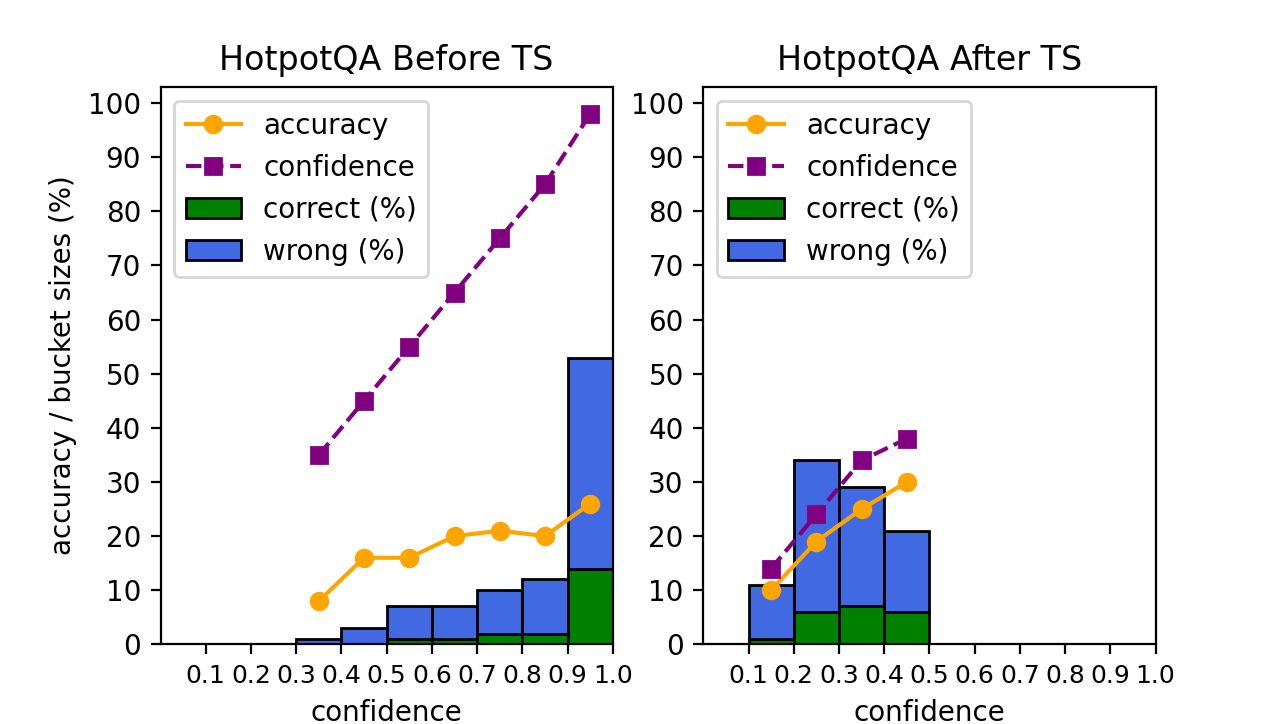}
\caption{Distribution of predictions on \abr{HotpotQA} in an \abr{ood}
  setting. We put predictions within the same confidence range into
  the same bucket (10 fixed-range buckets), compute the average
  confidence and accuracy within each bucket.
The x-axis represents the confidence range of each bucket, the y-axis
represents the average answer accuracy for the dashed line plot and
represents the relative bucket sizes for the histogram.
Before calibration, most predictions have overly high confidence. After temperature scaling, all predictions' confidence values are scaled to become closer to the overall answer accuracy (24.5). 
Both correct (green bars) and wrong predictions (blue bars)
are mixed in the same buckets, making them hard to distinguish.
% \jbgcomment{I think this caption could be clearer.  This figure is trying to do the job of explaining ECE and motivating our metric.  The reader probably doesn't understand what a ``bucketing distribution'' is.  Perhaps call this something like ``If you bucket each prediction from a model such that \dots''}
}
\label{fig:ece_illustration}
 \end{figure}

%\begin{comment}
%\chen{I feel the following paragraph is a bit disconnected, so leave %the comment here. } \sewon{I agree with this comment. I think the flow is good even if we comment out the following paragraph.}
%\chenglei{Who wrote the following para? Commenting out for now, lmk if you want to put it back.}
%
%A common problem in human--computer interactions is whether a user should accept a system's output.
%
%Indeed, this core problem is the foundation of interpretable machine learning.
%
%However, neural models struggle with even the basic task of creating a reliable measure of how confident its answer is.
%\end{comment}
%

% \jbgcomment{Is ``revisit'' the right word?  Perhaps reexamine, reconsider, etc.}
We re-examine calibration and apply it to a complex task with
real-world applications: open-domain question answering~\cite[\abr{odqa};][]{chen2017reading}.
The task takes an input question, retrieves evidence passages from a large
corpus such as Wikipedia, and then returns an answer string.
%\sewon{Minor: knowledge base -> corpus?}
%
%\sewon{I propose to cut the following paragraph, or only keep the first and the last sentence and merge with the previous paragraph.}
%\chen{Agree this paragraph can be shortened.}
%\jbgcomment{Moved some of the ideas to first paragraph}
% We focus on ODQA because it is a real-world example of a user forced to evaluate whether or not to trust a computer's prediction.
%
% From a user's perspective, a better confidence score could help a user understand whether to accept an answer or to dig deeper and verify an answer.
%
% From a provider's perspective, better calibration could prevent embarrassing wrong answers like answering how many legs a horse has with ``six''. %
%
%From an engineering perspective, 
%\abr{odqa} has two components (which we review in detail in section~\ref{subsec:base-model}): a passage retriever, and a reader for selecting the answer span. This makes the task format drastically unlike simple multi-class classification, and makes adapting existing calibration techniques a non-trivial challenge. 
Unlike classification, \abr{odqa} is a pipeline with multiple
components: a passage retriever followed by a reader.
This complexity is both typical of modern machine learning systems and
poses additional challenges.
%
% Third, the output answer space is complex, which includes single answer, multiple answers, and no answer. 
%Therefore, well-calibrated ODQA approaches have to consider both model components and complex labels, which lead to additional challenges. 
%
We explore adapting the calibration methods for the retriever-reader
\abr{odqa} models on both in-domain and out-of-domain (\abr{ood})
settings.
%We start with the conventionally used calibration metric, 
%expected calibration error (\abr{ece}).
% , which 
% buckets the model predictions into confidence intervals and computes
% weighted average of the 
% discrepancy between each bin’s expected accuracy and confidence. 
%After adapting and applying temperature scaling on \abr{odqa}, models get very low \abr{ece} scores (Section~\ref{sec:calibrate_qa}).
% \jbgcomment{Temperature scaling should be introduced a little more more smoothly}
Using the commonly used temperature scaling (\abr{ts}) method on
\abr{odqa}, models get lower \abr{ece}, similar to previous findings
on multi-class classification
tasks~\cite{guo2017calibration,Desai2020CalibrationOP}.
% (Section~\ref{sec:calibrate_qa}).
%
%

% \chen{I made relatively big changes for the following three paragraphs, let me know if you like them}

However, we argue that low \abr{ece} does not correspond to useful
calibration: in fact, it under-estimates the true calibration errors
due to its bucketing mechanism.
% 
% \jbgcomment{Bucketing needs to be introduced a little more cleanly or delayed until it can be fully explained; see the introducing concepts section in ``Toward Style and Grace''}
% \chenglei{Wait that's actually a book? I need to write that sometime...}
\abr{ece} measures the difference between the confidence and expected
accuracy by splitting the confidence values into buckets, taking an
average confidence and an average accuracy of each bucket and
marginalizing over their differences.
%. To do so, the standard practice is to split the confidence values into buckets so that we can compute the average confidence and accuracy of each bucket and marginalize over their differences. 
However, this allows models with middling confidence to win on the
\abr{ece} metric.
For instance, temperature scaling assigns all predictions in the range
$[0.1, 0.5)$ on \abr{HotpotQA} (Figure~\ref{fig:ece_illustration}),
which is not useful for users separating correct and wrong predictions
because the confidence values are all in a similar range.
% \jbgcomment{Say directly why it's not useful.}
%
%In temperature scaling, the temperature value is tuned on the development set to scale confidence values closer to the accuracy,
%therefore causing most predictions to be assigned similar confidence values. For instance, as shown in Figure~\ref{fig:ece_illustration}, all predictions (including correct and wrong ones) have confidence values in the range $[0.1, 0.5)$ after temperature scaling on \abr{HotpotQA}. 
%From a user-centric view, such similar confidence across all examples 
%does not help users to 
%distinguish correct and wrong predictions. In the case of Figure~\ref{fig:ece_illustration}, users cannot decide which predictions are trustworthy since all predictions have confidence lower than 0.5 after calibration.
%, which is the end goal for useful application. 
%which does not help a user trying to decide to
%trust a system's prediction; 
Moreover, the bucketing mechanism causes a \emph{cancellation effect}
where over-confident and under-confident predictions are bucketed
together and averaged out, hiding the instance-level calibration
errors.

We propose Macro-average Calibration Error (\metric{}) as an
alternative metric that directly focuses on distinguishing correct
from wrong predictions (Section~\ref{sec:macroce}).
% \jbgcomment{Forward point to definition}
%
\metric\ removes the bucketing mechanism and sums calibration error at
the \emph{instance} level.
% , in order to avoid the cancellation effect.
It also takes equal consideration of correct and wrong predictions
through macro-averaging in order to be insensitive to the accuracy
level (\textit{e.g.}, when the accuracy is very low, simply lowering
confidence on all predictions would lower \abr{ece}, but not
\metric{}).
%
% \jbgcomment{Replace ``we show'' with a section}
% We show that 
\metric{} is insensitive to accuracy shifts, successfully satisfying
the desiderata for a stable calibration
metric~\citep{nixon2019measuring}.  We also show that this metric
flips the conclusion from the previous section based on
\abr{ece}---four existing calibration methods, including temperature
scaling---the \abr{ece} winner---do not lead to improvements in
\metric{} (Section~\ref{sec:exp-new-metrics}).

%\sewon{Revised version ends here}

%\sewon{I have an impression that this new calibration method should be highlighted more.}
%\chen{Agree, and I think we should probably put this method beforehand, I
%  propose to introduce it before experimenting with existing methods.}

% \jbgcomment{I think a likely criticism is that we created a metric and surprise, surprise our new metric does better on it.  So it would be better to have a clearer motivation of where this calibration comes from (both here in short form and later).}
% \chen{I agree, add more intuitions in the following paragraph.}

%% cz0620 comment out the following sentence, it's duplicated. 
%Given the better metric, 
%a wide range of existing calibration approaches do not actually improve calibration, contrary to previous findings. %beliefs. 
%Towards better calibration, we aim to find additional cues to expose the model's confidence. 
% we find that existing calibration approaches rely heavily on the final model predictions, while a known issue is that models easily make incorrect predictions with high confidence. To mitigate this issue, 
%Towards this end, we propose to exploit the model's prediction consistency throughout the training trajectory as the indicator for confidence. 
To address this shortcoming, we propose a new method, \name{}, which
tracks whether the model makes \emph{consistent} correct predictions
over different checkpoints during training.
The intuition is that if the same correct prediction is consistent
throughout the training trajectory, then it could serve as a strong
sign that the model is confident about the prediction.
%Based on this intuition, we propose a new method
%(\name{}) that
%
%
%Thus, to address this gap we introduce a calibration method that improves
%MacroCE.
%
% rather than looks at final predictions, but 
%tracks whether the model makes 
%\emph{consistent} correct predictions over different checkpoints during training.
\name{} significantly improves \metric{} on both in-domain and
\abr{ood} evaluation (Section~\ref{sec:new-eval}), including when
downstream users must validate model predictions (Section~\ref{sec:new-eval}).

%we use the intuition that
%whether the model answers consistently during training is a strong cue for the
%prediction confidence (Section~\ref{sec:better}).

%\sewon{Proposal: have a bullet-point contribution list? Since there's lots of things going on, summarizing contributions might be useful. I put my proposal below.}

\jbgcomment{I'd like this more if this were more parallel.  Having
  some of the bullets have multiple sentences breaks up the flow}
In summary, our contributions are:
\begin{enumerate*}
    \item We thoroughly study calibration in the \abr{odqa} setting, 
    an under-explored real-world problem involving complex pipelines.
    We find that existing calibration methods like \abr{ts} achieve very low \abr{ece}; however, it does not capture if the model effectively distinguishes correct and wrong predictions.
    \item We propose a better metric \metric{}, which captures models' ability to distinguish correct and wrong predictions without being sensitive to the absolute model accuracy. Four different calibration methods known to be effective on \abr{ece} are ineffective on \metric{}.
    \item We introduce a new calibration method (\name) that uses the consistency of model predictions to estimate confidence, significantly reducing \metric{} and outperforming all previous baselines.
    \item Together, we provide an alternative view of calibration, re-evaluation of existing calibration methods and a proposal of a better calibration method based on our new viewpoint.
\end{enumerate*}

\section{Background}
\label{sec:background}

% \jbgcomment{Again, don't use bucketing until you can define it.}
This section reviews the existing
% bucketing-based 
calibration framework, the associated \abr{ece} metric, and the
commonly used temperature scaling method that effectively optimizes the \abr{ece}
metric.

\subsection{Bucketing-based Calibration and ECE}
\label{sec:bucket_ece}

% \jbgcomment{Too big of a cite dump: break up with individual contributions or pick the most representative}
% \chen{fixed}

Under the existing calibration framework, a model is
``perfectly calibrated'' if the prediction probability (i.e.,
confidence) reflects the ground truth
likelihood~\citep{NiculescuMizil2005PredictingGP}.
Specifically, given the input $x$, the ground truth $y$ and the
prediction $\hat{y}$, the perfectly calibrated confidence
$\mathrm{Conf}(x, \hat{y})$ will satisfy:
$\forall p \in [0, 1], P (\hat{y}=y \g \mathrm{Conf}(x, \hat{y})= p) = p$.

%\paragraph{Expected calibration error (ECE)}
%In order to practically measure the calibration error (i.e., how much does the confidence deviate from the expected accuracy), we follow~\citet{guo2017calibration} and use expected calibration error (ECE) as the metric.

% \jbgcomment{Break up the equation if we have room then use words to talk about accuracy and confidence}
% \chen{edited}

Prior work~\cite{guo2017calibration} evaluates calibration with \textbf{Expected
Calibration Error} (\textbf{\abr{ece}}), where \textit{N} model predictions are bucketed into~$M$ bins and predictions within the same confidence range are put into the same bucket. 
%
% \jbgcomment{``based on'' is vague.  Give a more precise intuition.}
%
Let~$B_m$ be the $m$-th bin of $(x,
y, \hat{y})$ triples, the accuracy $\mathrm{Acc}(B_m)$ measures how many instances in the bin are correct, 
% \sewon{Random suggestion: since we never refer back to the equation, what about removing the equation numbers by using `equation*' instead of `equation'?}
\begin{equation*}
    \mathrm{Acc}(B_m) =\frac{1}{|B_m|}\sum_{i=1}^{|B_m|}\mathbb{I}(y=\hat{y}), \\
\end{equation*}
where $\mathrm{|B_m|}$ is the number of examples in the $m$-th bin, and $\mathrm{Conf}(B_m)$ computes the average confidence in the bin, 
\begin{equation*}
    \mathrm{Conf}(B_m) =\frac{1}{|B_m|}\sum_{i=1}^{|B_m|}\mathrm{Conf}(x, \hat{y}). \\
\end{equation*}
Finally \abr{ece} measures the difference in expectation between confidence and accuracy over all bins,
\begin{equation*}
    \mathrm{ECE} =\sum^{M}_{m=1} \frac{|B_m|}{N} |\mathrm{Acc}(B_m) - \mathrm{Conf}(B_m)|. \\
\end{equation*}
%
%\begin{align*}
%    %\mathrm{Acc}(B_m)&=\frac{1}{|B_m|}\sum_{i=1}^{|B_m|}\mathbb{I}(y=\hat{y}), \\
%    \mathrm{Conf}(B_m)&=\frac{1}{|B_m|}\sum_{i=1}^{|B_m|}\mathrm{Conf}(x, \hat{y}), \\
%    \mathrm{ECE}&=\sum^{M}_{m=1} \frac{|B_m|}{N} |\mathrm{Acc}(B_m) - \mathrm{Conf}(B_m)|,
%\end{align*}
%$$\mathrm{ECE} = \sum^{M}_{m=1} \frac{|B_m|}{n} |\mathrm{Acc}(B_m) - \mathrm{Conf}(B_m)|,$$ where $\mathrm{Acc}(B_m)=\frac{1}{|B_m|}\Sigma_{i=1}^{B_m}\mathbb{I}(y=\hat{y})$ and $\mathrm{Conf}(B_m)=\frac{1}{|B_m|}\Sigma_{i=1}^{B_m}\mathrm{Conf}(x, \hat{y})$.
%
%where \textit{N} is the total number of examples and $\mathrm{|B_m|}$
%is the number of examples in the $m$-th bin.
%
Most work uses equal-width buckets: 
% (\textbf{interval-based ECE})
a triple $(x, y, \hat{y})$ with
$\frac{m}{M} \leq \mathrm{Conf}(x, \hat{y}) \leq \frac{m+1}{M}$ is
assigned to the $m$-th bin. \citet{minderer2021revisiting} and
\citet{Nguyen2015PosteriorCA} also used equal-mass binning:
% (\textbf{density-based ECE}): 
predictions are sorted by their confidence values and $\frac{N}{M}$
triples are assigned to each bin.
We find little difference between equal-width and equal-mass binning
\abr{ece} results (Table~\ref{tab:temp_scale_additional}), and so we will use the more
common equal-width binning in the rest of our experiments.
%We refer to the ECE metrics computed under these two binning schemes as \textbf{interval-based ECE} and \textbf{density-based ECE} respectively. 

% Both methods pre-define the number of bins $M$, and different choices of $M$ could result in different ECE results~\cite{minderer2021revisiting}.
% has reported that the ECE value may significantly vary depending on the choice of $M$.

%\chen{Maybe we should have stronger justification about why we use TS for main experiments}

\subsection{Temperature Scaling}

% \sewon{I think we have two options here: explain extractive QA and
% how the QA model works in general, which will give the complete
% picture of how temperature scaling works, or just talk about the
% simplest form of temperature scaling and say we will say more about
% how we apply temperature scaling for QA in Section 3. I drafted a
% paragraph assuming the second option, but unsure if it is the best
% option -- let me know.}

% \chen{After reading through section 2 and 3, what about in section 2
% (background), we talk about calibration (definition, ECE,
% temperature scaling, and classification), and open-domain QA. Then
% in section 3, start with what's wrong with ECE, and \textbf{why it's
% important in QA}, the reason is that in Computer vision, previous
% methods already mention ECE's issue, but we would like to claim this
% issue is more noteworthy in QA. Then TS for QA.
% } \paragraph{Temperature scaling}

% \jbgcomment{Again, break up cite dump}

% \jbgcomment{This is assuming too much prior familiarity with temperature scaling.  Think about how you'd explain it to someone for the very first time.  What is it scaling and why?  I tried to take a first pass, but could probably use some fine-tuning.}

% \jbgcomment{Removed this since it didn't fit for a friendly introduction; perhaps move to related work? Temperature scaling is the
% multi-class extension of Platt scaling~\citep{platt1999probabilistic}}

% \jbgcomment{``candidate output'' is never defined}

% \chen{I prefer using labels instead}

% \chen{did a quick pass, fixed these comments}

Without calibration, the confidence is often too high (or less
commonly, too low): it thus needs to be \emph{scaled} up or down.
A widely-used calibration method is \textbf{temperature
scaling}~\citep{guo2017calibration},
which uses a single scalar parameter called the temperature~$\tau$ to scale
the confidence.
% \footnote{Note that there are also other calibration approaches such
% as feature-based classification, we include experiments on those
% methods in Section~\ref{sec:re-eval}.}
% \jbgcomment{I'm not sure I like ``label'' set, as it doesn't work well with the QA setting.  Can we pick a better term that would be more intuitive?}
The temperature value is optimized
on the dev set. Given the set of candidate answers $\mathcal{C}$ and the logit value
$\mathbf{z} \in \mathbb{R}^{|\mathcal{C}|}$ associated with the
prediction $\hat{y}$, the confidence for the prediction $\hat{y}$ that
is the $j$-th label in $\mathcal{C}$ is: 
\begin{equation*}
    \mathrm{Softmax}\left(\frac{\mathbf{z}}{\tau}\right)_j.
\end{equation*}
For classification, the temperature scalar~$\tau$ is tuned to optimize
negative log likelihood (\abr{nll}) on the dev set.
% ~\footnote{For \abr{odqa}, we tune $\tau$ to optimize dev set \abr{ece} since it is not multi-class classification.}
%while for
%open-domain question answering, the number of correct answers in the
%candidate set $\mathcal{C}$ varies (zero, one, or more). Hence, we
%optimize development set \abr{ece} instead of \abr{nll}.
%% cz0516: remove the following sentence, seems duplicated 
%Temperature values larger  than 1 would lower the  confidence while temperature values smaller than 1 would increase the confidence. A temperature value of 1 is equivalent to not doing any calibration.
Temperature scaling only changes the confidence---\emph{not} the
predictions---so the model's accuracy remains the same.
%
%We follow~\citet{Desai2020CalibrationOP} to search optimal temperature scalars by 
%iterating over the range of [0.01, 10.00] with an interval of 0.01.

% \jbgcomment{The bit about QA seems premature, that's fully discussed in the next section.  Either cut or make it clearer that it's a transition to the next section.}
% \chen{agreed, moved it to next section}
%\begin{comment}
%In the rest of this section, we use the temperature scaling as a motivating example; details on applying the temperature scaling to the open-domain QA model as well we other calibration methods are provided in Section~\ref{sec:re-eval}.
%\end{comment}

%\begin{comment}
%\sewon{Original version:}

%The conventional temperature scaling method searches for a temperature value that optimizes the negative log likelihood on the dev set. 

%However, in the context of extractive question answering, the task is not really a multi-class classification problem and it is common that a question has no correct predictions. Under such cases, the NLL loss would be zero, making the temperature search ignore the no-answer questions. To avoid such cases, we instead search for a temperature value that optimizes the interval-based ECE on the dev set. We then apply the found temperature value on the test set with the same number of buckets. 

%Also note that for all temperature search, we search within the range of (0, 10] with 0.01 interval, same as~\cite{Desai2020CalibrationOP}. 
%\end{comment}

\section{Calibration in Open-Domain Question Answering}
\label{sec:calibrate_qa}

This section adapts the bucketing-based calibration framework for
multi-class classification to \abr{odqa} and evaluates
this calibration method on multiple \abr{qa} benchmarks, both in- and
out-of-domain.

\subsection{The ODQA Model}\label{subsec:base-model}

We use the model from \citet{karpukhin2020dense}, consisting of retrieval and reader components.
%following the retrieve-and-read pipeline~\citep{chen2017reading}.
%
The retrieval model is a dual encoder that computes the vector
representation of the question and each Wikipedia passage, and returns
the top-$K$ passages with the highest inner product scores between the
question vector and the passage vector.
The reader model is a \abr{bert}-based~\citep{devlin-19} span extractor.
Given the concatenation of the question and each retrieved passage, it
returns three logit values, representing the passage selection score,
the start position score and the end position score
respectively. These three logits are produced by three different
classification heads on top of the final \abr{bert} representations. More
precisely,
\begin{align*}
    \mathbf{H}_i &= \mathrm{BERT}(q,p_i) \in \mathbb{R}^{h \times L}, \\
    z^\mathrm{psg}(i) &= \left(\mathbf{H}_i\right)_{\texttt{[CLS]}}\mathbf{w}_\mathrm{psg} \in \mathbb{R}, \\ 
    z^{\mathrm{start}}(i, s) &= \left(\mathbf{H}_i\mathbf{w}^\mathrm{start}\right)_s \in \mathbb{R}, \\
    z^{\mathrm{end}}(i, e) &= \left(\mathbf{H}_i\mathbf{w}^\mathrm{end}\right)_e \in \mathbb{R},
\end{align*} 
where $\mathbf{w}_\mathrm{psg}, \mathbf{w}^\mathrm{start}, \mathbf{w}^\mathrm{end} \in \mathbb{R}^h$ are trainable parameters.

% \jbgcomment{would be better to integrate these better into the previous paragraph.}

\subsection{Temperature Scaling For ODQA}\label{subsec:base-calibration}
%\sewon{This section is much clearer now, thank you for revising!!}

% Unlike classification problem, where the model directly outputs the probability of each class, in ODQA, the score of each candidate span relies on both passage and span scores. Therefore, as an approximation, 
The formulation of \abr{odqa} is unlike conventional multi-class
classification since it involves both the retriever and reader,
leaving the question of what we should base the confidence score on.
To adapt temperature scaling on \abr{odqa}, we take the set of top
span predictions as our candidate set $\mathcal{C}$.
%
% \jbgcomment{is ``label set'' the best name?}
%
Specifically,
we compute the raw score for each candidate span and then apply
softmax over $\mathcal{C}$ to convert the raw span scores into
probabilistic confidence values.  We explore two possible
implementations: \textbf{\textit{Joint Calibration}} considers both
passage and span scores; \textbf{\textit{Pipeline Calibration}}
selects highest scored passage first, then calibrates on span scores
only.

\paragraph{Joint Calibration.}
Given the top $k=10$ retrieved passages for each question and for
each passage's top $n=10$ spans, we have an answer set of $n \times k
= 100$ spans per question. We score each candidate by adding its passage, span start and span end score:
\begin{equation*}
  z^{\mathrm{start}}(\hat{i}, s) + z^{\mathrm{end}}(\hat{i}, e) + z^\mathrm{psg}(i).
\end{equation*}
We then apply temperature scaling to the predicted logits and the confidence becomes:
\begin{equation*}
\tiny
  \mathrm{Softmax}_{(i, s, e) \in \mathcal{C}}\left(\frac{z^\mathrm{psg}(i)+z^{\mathrm{start}}(i,s)+z^{\mathrm{end}}(i,e)}{\tau}\right)
\end{equation*}
Note that for
\abr{odqa}, the number of correct answers in the
candidate set $\mathcal{C}$ varies (zero, one, or more). Hence, the temperature scalar $\tau$
is tuned to optimize dev set \abr{ece} instead of \abr{nll}.

% \jbgcomment{These equations would be more effective with longer
%   English description of the match.  Remind the reader what the inputs
%   are and what $\tau$ does to that input.}
% \chen{edited}
  
% \jbgcomment{Actually contrast these two approaches.  Point out what's different.}
% \chen{add it at the beginning of the section}

\paragraph{Pipeline Calibration.}

We choose the passage with the highest passage selection score $i_\mathrm{max} = \mathrm{argmax}_{1 \leq i \leq K}z^\mathrm{psg}(i)$ and then define the span score $S(s, e, i)$ as
\begin{equation*}
    \left(z^{\mathrm{start}}(\hat{i}, s) + z^{\mathrm{end}}(\hat{i}, e)\right) \mathbb{I}[i=i_\mathrm{max}].
\end{equation*}
In this case, we only keep the top $n=10$ spans from the top passage for each question. Like Joint Calibration, we apply temperature scaling to the predicted span logits and the confidence is:
\begin{equation*}
    \mathrm{Softmax}_{(i, s, e) \in \mathcal{C}}\left(\frac{z^{\mathrm{start}}(i, s)+z^{\mathrm{end}}(i, e)}{\tau}\right).
\end{equation*}

\begin{table}[t]
\small
\begin{center}
\setlength{\tabcolsep}{1.5mm}{
\begin{tabular}{ l c @{\hspace{5\tabcolsep}} cccc }
    %& \multicolumn{2}{c}{$\overbrace{\phantom{~~~~~}}^{\text{Section~\ref{sec:exp}}}$} & \multicolumn{2}{c}{$\overbrace{\phantom{~~~~~~~~}}^{\text{Section~\ref{sec:exp-new-metrics}}}$} \\
    && \multicolumn{2}{c}{Section~\ref{sec:calibrate_qa}} & \multicolumn{2}{c}{Section~\ref{sec:exp-new-metrics}} \\
 \toprule
    Model & TS & EM$_{\uparrow}$ & ECE$_{\downarrow}$ & ICE$_{\downarrow}$ & \metric{}$_{\downarrow}$ \\ 
 \midrule
    \multicolumn{5}{l}{\textbf{\em NQ}} \\
    Joint & -  & 32.9 & 27.1  & 47.8 &  43.6 \\
    Joint & \checkmark & 32.9 & 4.0  & \textbf{37.4} & \textbf{42.5} \\
    Pipeline & - & 34.1 & 48.2  & 55.1 & 44.2 \\
    Pipeline & \checkmark & 34.1 & \textbf{2.7} & 39.7 & 44.4 \\
 \midrule
    \multicolumn{5}{l}{\textbf{\em NQ $\rightarrow$ \abr{HotpotQA}}} \\
    Joint & - & 24.9 & 41.0  & 54.7 & 45.7 \\
    Joint & \checkmark & 24.9 & 12.5  & 40.3 & \textbf{45.5} \\
    Pipeline & - & 22.6 & 59.6  & 65.9 & 47.4 \\
    Pipeline & \checkmark & 22.6 & \textbf{8.4} &  \textbf{37.9} & 47.7 \\ 
 \midrule
    \multicolumn{5}{l}{\textbf{\em NQ $\rightarrow$ \abr{TriviaQA}}} \\
    Joint & - & 33.6 & 25.4  & 48.6 & 45.1 \\
    Joint & \checkmark & 33.6 & 6.4  & \textbf{38.4} & 44.3 \\
    Pipeline & - & 34.2 & 48.2  & 54.5 & \textbf{43.7} \\
    Pipeline & \checkmark & 34.2 & \textbf{6.1} &  39.2 & 44.6 \\
    \midrule
    \multicolumn{5}{l}{\textbf{\em NQ $\rightarrow$ \abr{SQuAD}}} \\
    Joint & - & 12.4 & 41.7  & 48.5 & \textbf{39.5} \\
    Joint & \checkmark & 12.4 & \textbf{12.4} &  \textbf{26.6} & 39.7 \\
    Pipeline & - & 12.2 & 62.7 & 65.1 & 41.4 \\
    Pipeline & \checkmark & 12.2 & 13.5 & 29.1 & 43.9 \\
 \bottomrule
\end{tabular}}
 \caption{
    In-domain and \abr{ood} calibration results. \textit{Joint} and \textit{Pipeline} refer to whether the candidate set consists of top answer candidates from all top-10 retrieved passages or just the top-1 retrieved passage. All numbers are multiplied by 100 for better readability throughout the paper.
    \abr{em}: higher is better. Calibration errors: lower is better. Best calibration result in each group is in bold. Across all settings, temperature scaling significantly improves \abr{ece} but not \metric{}, highlighting the difference between these calibration metrics. Also, \abr{ood} incurs higher calibration errors. 
    % \jbgcomment{We might want to think about if this might be better as a figure, particularly since that might work better for the presentation (since we'd only want to focus on a few results)}
 }
 \label{tab:temp_scale}
\end{center}
\end{table}

\subsection{Temperature Scaling Results}\label{subsec:base-exp-setup}

% \jbgcomment{This could use a clearer section title}

%\chenglei{Will also put some other implementation details here.}
%We introduce the QA datasets that we will be using is this paper: NQ, SQuAD, TriviaQA, HotpotQA. 

We experiment the above temperature scaling methods on both in-domain
and \abr{ood} settings, since
\citet{Desai2020CalibrationOP,jiang2021can} argue that \abr{ood}
calibration is more challenging than in-domain calibration.
%We use four QA datasets: \textbf{Natural Questions (NQ)}~\citep{kwiatkowski-19}, \textbf{SQuAD}~\citep{rajpurkar-16}, \textbf{TriviaQA}~\citep{joshi-17} and \textbf{HotpotQA}~\citep{yang2018hotpotqa}.\footnote{We include dataset details in appendix.}
We use \textbf{\abr{NaturalQuestions}}~\cite[\abr{nq}]{kwiatkowski-19} as the in-domain dataset, and
\textbf{\abr{SQuAD}}~\citep{rajpurkar-16},
\textbf{\abr{TriviaQA}}~\citep{joshi-17} and
\textbf{\abr{HotpotQA}}~\citep{yang2018hotpotqa} as the out-of-domain
datasets.\footnote{More dataset details are in appendix~\ref{sec:datasets}. }
%
%\sewon{I think I wrote these dataset descriptions, but now think we can cut them if we need to save space, since we aren't looking into details of each dataset anyways.}
%
%
We tune hyper-parameters  on the in-domain \abr{nq} dev set.
%For temperature scaling, we tune the temperature scalar on the NQ development set, evaluate on NQ test set (in-domain) and three other QA datasets (OOD).
We report exact match (\abr{em}) for answer accuracy
% (note that temperature scaling does not impact the EM score),
and \abr{ece} for calibration results. 
%Note that temperature scaling does not impact the EM score.

% \paragraph{Implementation details.}
% We follow \citet{karpukhin2020dense} in using the plain text portion of the English Wikipedia dump from 12/20/2018. We use $K=10$.

%\paragraph{Candidate Generation}
% When computing the candidate set $\mathcal{C}$, enumerating over all possible spans of $K$ passages is expensive. We therefore approximate $\mathcal{C}$ by taking the top 10 predictions from each of $K$ passage, resulting in $|\mathcal{C}|=100$ with $K=10$.

%We first obtain a list of candidate answers for each question. We explored two possibilities: 1) We keep the top 10 predictions from each of top 10 retrieved passages. This results in a total of 100 candidate predictions per question. We name this approach \textit{joint} calibration. 2) We only choose from the top 10 predictions from the passage with the highest passage selection score. This results in a candidate pool of size 10 instead of 100. We name this approach \textit{pipeline} calibration.  By default we use the pipeline approach as we found it to produce slightly better calibration results.

% \subsection{Results}

% \subsection{Results}\label{subsec:base-results}

% \chenglei{Will also add tables for the OOD datasets. Will re-write this section after I fill in these tables.}

Without calibration, both joint and pipeline approaches have high
\abr{ece} scores, and the pipeline approach incurs higher
out-of-the-box calibration error (Table~\ref{tab:temp_scale}).
Applying temperature scaling significantly lowers \abr{ece} in all
cases, including both in-domain and \abr{ood} settings.
As expected, \abr{ood} settings incur higher \abr{ece} than the
in-domain setting even after calibration.
However, in the next section, we challenge this ``success'' by
re-examining the bucketing mechanism in \abr{ece} computation.

\section{Flaws in ECE and Better Alternatives}
\label{sec:exp-new-metrics}
%In the previous section, we showed that adapting temperature scaling on ODQA improves ECE significantly. However, in this section, we challenge this success by re-examining the bucketing mechanism in ECE computation. 
%
% Despite current approaches work well, we find the evaluate metric, ECE, has major flaws.
%the popularity of ECE (in particular the interval-based ECE) as the calibration metric, we find major flaws of ECE.
This section takes a closer look at the model accuracy and confidence,
and illustrate how \abr{ece} is misleading in evaluating model
calibration. We provide complementary views of calibration and propose
a new calibration metric as an alternative.

%This section uses case studies to illustrate the flaws
%of \abr{ece}, provides complementary views of calibration, and
%details our new calibration metric.
%\sewon{Proposal: \\
%In this section, we take a closer look at the model accuracy and confidence, and illustrate how \abr{ece} is misleading in evaluating model calibration. We provide complementary views of calibration, and
%propose a new calibration metric as an alternative.}
%\sewon{I think this section can overall be more concise. I can take a pass once the figure is ready.}

% \chen{Do you think there are some duplicates in 4.1 and 4.2?}

\subsection{What's Wrong With ECE?}

% \chen{What about something like Gap between ECE and real calibration setting?}

% In this section, we use some illustrative examples to demonstrate the flaws of the ECE metric. In particular, we show that low ECE does not necessarily correspond to good calibration performance due to the bucketing mechanism. 

% We begin with a case study on \abr{ood} setting where we train
% \abr{dpr-bert} on NQ and evaluate on \hotpotqa{}.
We illustrate the \abr{ece} problem with a case study
on \abr{HotpotQA}; similar trends surface for other datasets
(Appendix~\ref{app:more_ece}).
The uncalibrated model is  over-confident (Figure~\ref{fig:ece_illustration}): the confidence is
higher than the accuracy.
% \jbgcomment{I'm seeing ``we observe'' and ``we find'' a lot.  This usually can be cut.  Run the style checker script.}
%
After temperature scaling, the accuracy and confidence converge,
reducing \abr{ece}.
However, this over-estimates the
effectiveness of temperature scaling for 
two reasons.
% \footnote{Our findings apply to both equal-width and
% equal-mass binning of \abr{ece}, although we use equal-width binning
% for illustration.\chen{why?}\sewon{+1. We don't have explanation for
% this...}}
%
%
%
%  \begin{figure}[t]
% \centering
% \includegraphics[width=0.50\textwidth]{figures/hotpotqa_uncalibrated.png}
% \caption{Uncalibrated results on HotpotQA. x-axis: confidence ranges of buckets. Solid line: average accuracy of each bucket. Dashed line: average confidence of each bucket. Histogram: relative sizes ($\%$) of each bucket.
% \sewon{Proposal: let's explicitly put the axis title? I think it'll be obvious for people familiar with calibration, but won't be obvious for others.}
% \sewon{Proposal: Combine Fig 2 and 3, have figures for NQ as well, so that there is one 2-column figure with four small subfigures? IMHO this is the most important figure in this paper.}
% \chen{I agree with Sewon's proposal, those four figures will look better, also could save some space. }
% }
% \label{fig:hotpotqa_uncalibrated}
% \end{figure}
%
% \begin{figure}[t]
% \centering
% \includegraphics[width=0.50\textwidth]{figures/hotpotqa_ts.png}
% \caption{Results on HotpotQA after temperature scaling. x-axis: confidence ranges of buckets. Solid line: average accuracy of each bucket. Dashed line: average confidence of each bucket. Histogram: relative sizes ($\%$) of each bucket.}
% \label{fig:hotpotqa_ts}
% \end{figure}
%
%
%
%We outline the reasons why such calibration result is not ideal and how the interval-based ECE is misleading:
First, \textbf{most instances are assigned similar confidence.}
%\sewon{-> Most instances assigned to 1--2 bins?}
All predictions have a confidence score between 0.1 and 0.5, not giving useful cues except that the model is  not confident on most predictions.
%After calibration, all predictions have low confidence (which match the relatively low accuracy).
%
%This is essentially conveying the message that all predictions are not confident, even for correct ones.
% Such calibrated predictions therefore do not allow users to identify correct predictions.  
% We argue that the low ECE under such case has little practical implication because it is not telling the difference between correct and wrong predictions.
%This would be fine if there are \emph{no} examples we should be confident about.
%
This is not ideal since, if there were examples we could trust or abstain, an ideal calibration
metric should recognize such scenarios and encourage the calibrator to
differentiate correct and wrong predictions.
Second, \textbf{bucketing causes cancellation effects, ignoring instance-level calibration error}. % thus underestimating the true calibration error.}
Many predictions are clustered in the same buckets. 
%(in the [0.1, 0.4] range in Figure~\ref{fig:ece_illustration}). 
As a result, there are many over-confident and under-confident
predictions in the same bucket 
% whose effects are cancelled out
% when they are being averaged.
and are averaged to become closer to the
average accuracy.~\footnote{In fact, these flaws apply to many other NLP tasks. An example on sentiment analysis is shown in Table~\ref{fig:more_ece_illustration_sst2}.}
%However, on the instance level, over- and under-confidence still exists for wrong and correct predictions. 
% For instance, when we evaluate the same temperature value on the subset where the question has no correct prediction (where the models tends to be overconfident), the ECE is 21.58; and when we evaluate on the subset where the question has multiple correct predictions (where the model tends to be underconfident), the ECE is 21.03.\chen{This example seems unnecessary to me} Both numbers are much larger than the ECE on the entire test set, illustrating the issue that the cancellation effect gives a misleadingly low ECE result. 
%A better calibration metric should avoid such effects caused by bucketing.

%We include more such examples on different test sets including both IID and OOD evaluations in the appendix. We find that these problems persist across all settings and the conventional interval-based ECE metric fails to reflect these issues. 

\paragraph{The Need for An Alternative View.}
% \sewon{I love the intuition in this section. Maybe we can consider merging it with Section 2.4?}
% \chen{Agree}
%
% \jbgcomment{``Expectation point of view'' is awkward phrasing}
The above issues are because \abr{ece} only measures the expectation
where the aim is to match the confidence with the expected accuracy.
However, this goal can be trivially achieved by simply outputting the
similar confidence for all predictions that match the expected
accuracy, as in the case of temperature scaling, which is not useful because 
% actual goal is to differentiate correct and wrong
% predictions.
%
%In practical settings, such calibration results are not particularly
% useful because 
users cannot easily use such confidence scores to make abstaining judgments.
% they just generally less trust the model.
%
Hence, we propose an alternative view of calibration where the goal
is to \textbf{maximally differentiate  correct and wrong predictions}.
We argue that achieving this goal would bring better practical values
for real use cases.  Toward this end, we propose a new calibration
metric that aligns closely with our objective.

% \sewon{Possible Q: this mainly explains equal-width binning. Why is equal-mass binning not good either?}
% \chen{Interesting, I did not see equal-mass binning mentioned in this section (but agree readers may have this question), what about we add a footnote explaining this problem?}
% \chenglei{I said at the end of 2.1 that equal-width and equal-mass have very similar trends which I think would account for this - same problems here apply for equal-mass as well.}

%\chen{Maybe expand the following subsection a bit more, will discuss it in the meeting.}

% \chen{Now it seems that we start with introducing both ICE and MarcoCE, does some experiments, and makes the decision that MarcoCE is better, i agree that this seems like a better flow}

\subsection{New Metric: \metric{}}
\label{sec:macroce}
We propose alternative metrics that remove the bucketing mechanism to prevent the above problems.
We consider two such metrics---\abr{ice} and \metric.
We evaluate their robustness to various distribution shifts, and propose to use \metric\ as the main metric.

%\chen{Add a footnote saying we only consider 0/1 prediction, but can extend to multi-class, and add sth in limitations section}

% \jbgcomment{Why aren't we using paragraph tags?}
% \sewon{I was told at some point not to use the paragraph tag if it is part of the sentence rather than a title, but I'm fine either way.}
% \jbgcomment{I think using paragraph is fine here; much better than this hacky workaround with vspace an noindent.}

\paragraph{Instance-level Calibration Error (\abr{ice})} accumulates
the calibrator error of each individual prediction and takes an
average.\footnote{This is equivalent to \abr{ece} with equal-mass
  binning under the condition that $M=N$ (i.e., the bucket size is
  always 1). } Formally,
%\sewon{Proposal: hint first that the MacroCE is the main metric? People might be confusing because the title says MacroCE and now we are introducing ICE.}
\begin{equation*}
    \mathrm{ICE} = \frac{1}{n}\sum^{n}_{i=1} |\mathbb{I}(y_i=\tilde{y}_i) - \mathrm{Conf}(x_i, \tilde{y}_i)|.
\end{equation*}
%
%Nonetheless, one problem of the \abr{ice} metric is that it is highly sensitive to the model accuracy (shown in Section~\ref{subsec:base-analysis}).
%
\abr{ice} is similar to the Brier Score~\cite{Brier1950VERIFICATIONOF}
except that we are marginalizing over the absolute difference between
accuracy and confidence of predictions, instead of squared
errors.\footnote{We prefer the L1 over L2 norm because L2 norm favors
  confidences in the middle ranges rather than the binary ends, which
  is less useful for a user trying to decide if an outcome is good or
  not.}

% \jbgcomment{It might worthwhile to talk about the Brier score; why aren't more people using it, etc. in the related work.}

%However, 
% as we will show in the next section, 
%the \abr{ice} metric has another important flaw: it is highly sensitive to the model accuracy. \sewon{Cite the vision paper that claims it shouldn't be sensitive to accuracy?} \todo{for chenglei}
%For instance, if most predictions are correct, applying a small temperature scalar to boost all confidence scores lowers the \abr{ice} score, but not penalizing the wrong predictions with high confidence scores, as the small proportion of wrong predictions would not contribute much to the overall metric. 

% \jbgcomment{I think the macro-averaging needs to be foregrounded saying something like: traditional measures focus on wrong cases because they're overconfident on errors.  Because we want this to explicitly help know when an answer is right, we need to focus on correct predictions as well.  We do this by explicitly computing \dots for correct predictions and then combine that into a single score by \dots}

While \abr{ice} and Brier Score prevent the issues brought by
bucketing, it incurs another issue: they can be easily dominated by
the majority label classes.
For example, if the model achieves high accuracy, always assigning a
high confidence can get low \abr{ice} and Brier Score because the
wrong predictions contribute very little to the overall calibration
error (we will empirically show this in the following
experiments). This is undesirable because even when the wrong
predictions are rare, mistrusting them can still cause severe harms to
users.
To address this, we additionally perform macro-averaging of the
calibration errors on correct and wrong predictions, and we name this
metric \metric{}.

\paragraph{Macro-average Calibration Error (\metric)} considers instance-level errors, but it takes equal consideration of correct and wrong predictions made by the model. Specifically, it calculates a macro-average over calibration errors on correct predictions and wrong predictions:
%\chen{The following equation maybe a bit confusing, the use of $p$ and $n$}\sewon{QQ: Isn't $n$ different from the ICE equation, and also different between pos and neg?}
%\chen{Agree, what about $n_{pos}$ and $n_{neg}$? does not look pretty, but leave it there}
%
%To overcome this issue, we propose to take equal consideration of both correct and wrong predictions with the \textbf{Macro-average Calibration Error (\metric)}: 
%\sewon{delete category? I have been thinking what category means was never clear.}
% \todo{Fix the following eqs, too many space, use align}
% \begin{align*}
%     \mathrm{Acc}(B_m)&=\frac{1}{|B_m|}\sum_{i=1}^{|B_m|}\mathbb{I}(y=\hat{y}), \\
%     \mathrm{Conf}(B_m)&=\frac{1}{|B_m|}\sum_{i=1}^{|B_m|}\mathrm{Conf}(x, \hat{y}), \\
%     \mathrm{ECE}&=\sum^{M}_{m=1} \frac{|B_m|}{N} |\mathrm{Acc}(B_m) - \mathrm{Conf}(B_m)|,
% \end{align*}
\begin{align*}
\mathrm{ICE_{pos}} &= \frac{1}{n_{p}} \sum^{n_{p}}_{i=1} (1 - \mathrm{Conf}(x_i, \tilde{y}_i)), \forall \tilde{y}_i = y_i, \nonumber \\
\mathrm{ICE_{neg}} &= \frac{1}{n_{n}} \sum^{n_{n}}_{i=1} ( \mathrm{Conf}(x_i, \tilde{y}_i) - 0), \forall \tilde{y}_i \neq y_i, \nonumber \\
\mathrm{MacroCE} &= \frac{1}{2} (\mathrm{ICE_{pos}} + \mathrm{ICE_{neg}}).
\end{align*}
Where $n_p$ and $n_n$ are the number correct and wrong
predictions.\footnote{The idea of macro-averaging was also referred to
  as class conditionality in \citet{nixon2019measuring} in the context
  of calibrating image classifiers, but an important distinction is
  that \metric{} removes bucketing.}
% \jbgcomment{Like the Brier score, I think class conditionality could be better discussed in the related work where it would be less distracting and you could expand a little bit more}
%To examine the appropriateness of the new metrics, we evaluate the robustness of these metrics under shift in accuracy and shift in data subpopulation. Ideally, a good calibration metric should be insensitive to these shifts and stably reveal models' calibration in all situations~\cite{nixon2019measuring}. 
%
Ideal calibration metrics should be insensitive to shifts in
accuracy~\citep{nixon2019measuring} and stably reveal models'
calibration in all situations. We examine the robustness of these metrics.

%In this way, $\mathrm{MacroCE}$ gives equal weight to positive (correct) and negative (wrong) predictions. We will demonstrate the advantage of %doing so in the next section. 

%In the following subsection, we provide a series of controlled experiments to highlight why \metric{} is more informative. 

%\subsection{Comparing Different Metrics Through Controlled Experiments}\label{subsec:base-analysis}

%We design three controlled experiments (NQ in-domain evaluation) to compare \abr{ece} and \metric. 
%These results would illustrate why MacroCE provides a better account oscf the actual calibration results. 

% For temperature scaling, we perform three sets of experiments for a thorough analysis on how TS works and use the results as a lens to better understand what different metrics capture.

 \begin{figure}[t]
 \centering
 \includegraphics[width=0.9\linewidth]{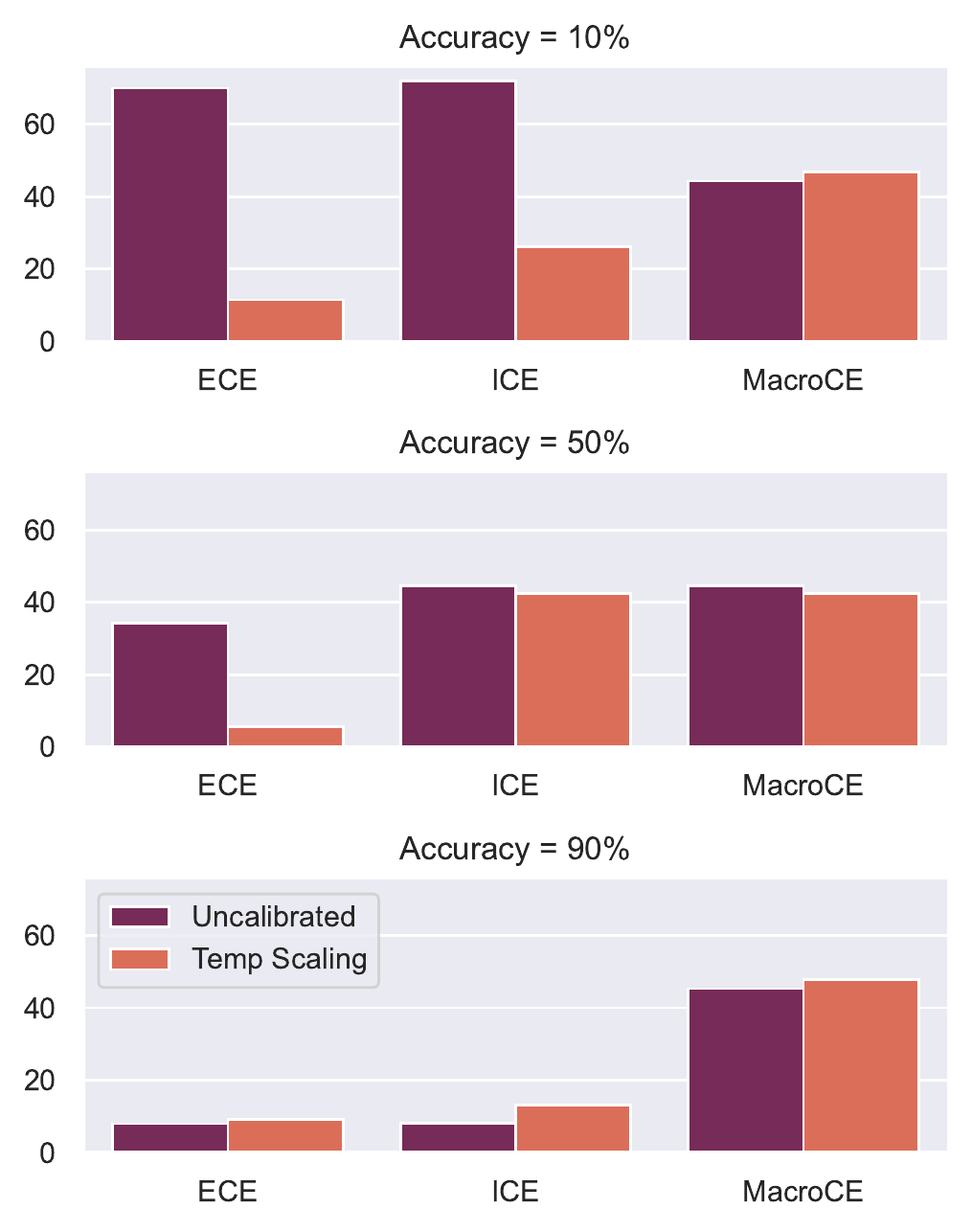}
\caption{
Calibration results where we re-sample the predictions to vary the model accuracy (10\%, 50\%, 90\%; same on both dev and test sets).
% ``Temp'' represents the temperature value tuned on the development set. 
Uncalibrated \abr{ece} and \abr{ice} results differ largely  at different accuracy (\textit{i.e.}, highly sensitive to accuracy), while \metric{} stays much more stable.
}
\label{tab:analysis2}
 \end{figure}

\paragraph{Temperature Scaling at Different Accuracy Levels}
%re-sample the dev and test set of NQ to
% \jbgcomment{``control'' is vague here.  Be more precise.}
We re-sample the data to very the model accuracy and examine the
effect of temperature scaling at different accuracy levels.
% An ideal metric should not be sensitive to the model accuracy. 
%
According to Table~\ref{tab:analysis2}, before calibration, the
\abr{ece} score decreases with higher model accuracy, since
%This is not because models are becoming better calibrated (since we are using the same model checkpoint), but rather, the 
higher accuracy matches the over-confidence predictions and gets
rewarded by low \abr{ece} score. Such finding also applies to
\abr{ice}, since the majority of predictions are correct, the impact
of negative predictions with over-confidence is marginal.
%metric  is correct predictions at $90\%$ accuracy. 
%as well. As a high accuracy ($90\%$), the high confidence matches the high accuracy. 
%However, the negative predictions have over-confidence but their impact is marginal to the overall ICE metric since the majority is correct predictions at $90\%$ accuracy. 
\metric{} results remain stable across all accuracy levels. As model
accuracy increases, \abr{ice}$_{\mathrm{pos}}$ decreases and
\abr{ice}$_{\mathrm{neg}}$ increases. \metric{} captures the trade-off
and implies the model remains poorly calibrated.\footnote{We also
present the numerical results in Table~\ref{tab:analysis2_app}.}

 \begin{figure}[t]
 \centering
 \includegraphics[width=0.9\linewidth]{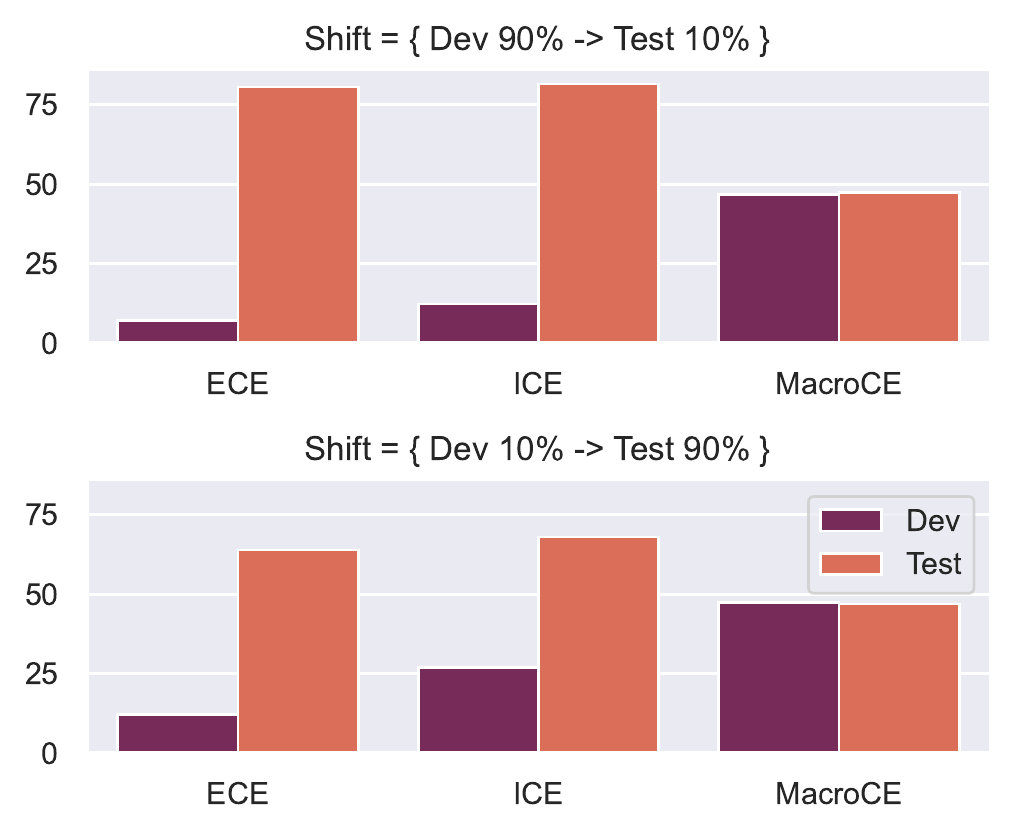}
\caption{
 Calibration results when training and test accuracy are different. In the first case, we tune the temperature value on a dev set with only 10\% correct predictions and a test set with 90\% correct predictions, and we reverse the setup in the second case. 
\abr{ece} and \abr{ice} change significantly under such accuracy shifts even though the underlying model is the same. In contrast, only \metric{} is stable under train-test accuracy shifts as desired.
}
\label{tab:analysis3}
 \end{figure}

% \begin{table}[t]
% \small
% \begin{center}
% \setlength{\tabcolsep}{2mm}{
% \begin{tabular}{ l c c c c }
%  \toprule
%   & Temp & ECE & ICE & \metric \\ 
%  \midrule
%     \multicolumn{5}{l}{\textbf{\em Development=90\% $\rightarrow$ Test=10\%}} \\
%     Development & 0.47 & 7.20 & 12.29 & 46.64 \\
%     Test & 0.47 & 80.54 & 81.49 & 47.39 \\
%  \midrule
%     \multicolumn{5}{l}{\textbf{\em Development=10\% $\rightarrow$ Test=90\%}} \\
%     Development & 10.00 & 12.23 & 26.88 & 47.48 \\
%     Test & 10.00 & 63.87 & 68.00 & 46.94 \\
%  \bottomrule
% \end{tabular}}
%  \caption{Calibration results when training and test accuracy are different. In the first case, we tune the temperature value on a development set with only 10\% correct predictions and a test set with 90\% correct predictions, and we reverse the setup in the second case. 
% \abr{ece} and \abr{ice} change significantly under such accuracy shifts even though the underlying model is the same. In contrast, only \metric{} is stable under train-test accuracy shifts as desired.}
%  \label{tab:analysis3}
% \end{center}
% \end{table}

%\chen{I think we should combine table 2 and 3 into one table.}

\paragraph{Temperature Scaling under Accuracy Shift}
We consider the  setting where there is a
large difference between the dev and test set accuracy.
In particular, we consider the cases where we: 1) tune temperature on
a development set with 90\% accuracy and evaluate on a test set with
10\% accuracy; and 2) tune temperature on a development set with 10\%
accuracy and evaluate on a test set with 90\% accuracy.
\abr{ece} and \abr{ice} are sensitive to the model accuracy but
\metric is not (Table~\ref{tab:analysis3}).
Temperature scaling
selects a low temperature scalar on the highly accurate development
set which does not transfer to the test set with low accuracy, which
indicates that temperature tuning does not hold against subpopulation
shift.
\metric{} reflects the poor calibration of the model on both cases,
insensitive to the  shift.\footnote{We also present the numerical results in Table~\ref{tab:analysis3_app}}.

% \chen{I added a footnote to indicate some limitations, can move it directly to the limitation section also}

Both experiments indicate that \metric{} is a more reliable calibration evaluation metric. 
We thus focus on \metric{} evaluation throughout the rest of the paper.
% \footnote{Note that while experiments show that \metric{} is a better metric over \abr{ece} and \abr{ice}, it's not the panacea for all use cases. 
% For example, in some applications other than question answering, the ideal confidence output might be $0.5$ to indicate the uncertainty of output, which is not encouraged by \metric{}. 
% We elaborate its limitation in Section~\ref{sec:limitation}.}
% In the next section, we examine whether other calibration techniques besides temperature scaling are effective in improving MacroCE.

%\chen{TODO: create a new paragraph saying \metric{} does not work for every case, and why for QA is more important}

\subsection{Re-Evaluating Calibration with New Metrics}

% \jbgcomment{Writing is clunky here.  But conclusion first, then back it up with evidence.}

Table~\ref{tab:temp_scale} compares calibration results under
\abr{ece} (with both equal-width and equal-width binning), \abr{ice}
and \metric{} results for all experiment settings described in
Section~\ref{subsec:base-exp-setup}.
Temperature scaling significantly improves \abr{ece} and \abr{ice}
with both joint and pipeline calibration, but it does not improve
\metric{}.
Moreover, \abr{ece} and \abr{ice} are very sensitive to accuracy
shifts. When transferring from \abr{nq} to HotpotQA and SQuAD where the
accuracy drops, there is significant increase in \abr{ece} and
\abr{ice} (before temperature scaling), but \metric{} stays high
despite such shifts.

% In comparison, \metric{} is not sensitive to model accuracy, and applying temperature scaling has no improvement on 
% \metric{}.
%we find the stark contrast that while temperature scaling brings significant improvement on ECE and ICE, it has no improvement on MacroCE. 
%Moreover, different metrics ranks different calibration methods differently, for example, for in-domain calibration, joint calibration gets worse ECE but better MacroCE over pipeline calibration.
%\sewon{Expand it more? Feel like this is an important discussion and ends really fast lol.}
%\chen{I added some take-aways, but guess we could add a few more.}

\section{Toward Better Calibration Methods}
\label{sec:new-eval}
% \chen{I think we need to expand, and a much better structured section.}
% \sewon{Restructured it, let me know what you think}
% \chen{Do we want to have \name{} as one subsection, and other approaches in another subsection?}
% \chenglei{Edited.}

% \chen{I think this section misses the intuition part. We need a few paragraphs saying why methods in section 5 do not work well, and the proposed approach works. }

% We first reflect the failure of existing calibration methods in section~\ref{sec:re-eval}.

% \jbgcomment{Maybe we shouldn't mix previous calibration with our new stuff?  Let's make the contribution clearer.}

This section reviews existing calibration methods other than
temperature scaling as baselines---including feature based classifier, neural
reranker and label smoothing (Section~\ref{subsec:baselines})---and introduces our new calibration
method based on model consistency throughout training trajectory
(\name{}; Section~\ref{subsec:conscal}).
All existing methods do not lower \metric{}, while \name{}
significantly improves \metric{} (Section~\ref{subsec:experiments}).

\begin{table*}[t]
\small
\setlength\tabcolsep{5pt}
\centering
\begin{tabular}{ lrrcrrc} 
& \multicolumn{3}{c}{\abr{iid} (\abr{nq)}} & \multicolumn{3}{c}{\abr{ood} (\abr{HotpotQA)}} \\
 \toprule
 Calibrator &  
 EM & ECE & \metric{} & EM & ECE & \metric{}
 \\
  \midrule
  No Calibration & 35.2 & 30.4 & 44.5 & 24.5 & 44.4 & 46.2 \\
  Binary Baseline & 35.2 & 38.0 & 53.2 & 24.5 & 38.6 & 44.4 \\
  Average Baseline & 35.2 & \textbf{2.0} & 50.0 & 24.5 & \textbf{11.3} & 50.0 \\
  \midrule
  Temperature Scaling &  35.2 & 4.7 & 42.5 &  24.5 & 13.7 & 45.6  \\
  %Feature-based (LR) & 37.3 & 61.5 & 49.0 & 21.4 & 76.6 & 48.9 \\
  Feature-based  & 36.5 & 52.3 & 45.0 & 21.8 & 62.4 & 46.9 \\
  %Feature-based (RF) & 36.4 & 47.4 & 44.8 & 24.6 & 53.9 & 46.9 \\
  Neural Reranker & 37.6 & 58.6 & 41.0 & 26.5 & 51.4 & 47.0 \\
  Label Smoothing & 36.1 & 29.4 & 45.6 & 23.6 & 44.7 & 46.8 \\
  Label Smoothing + TS  & 36.1 & 5.6 & 43.5 & 23.6 & 14.3 & 46.0 \\
  \midrule
  \name\ w/o Training Dynamics & 37.8 & 29.0 & 32.2  & 25.7 & 31.9 & 41.0 \\
  \name &  35.2 & 33.1 & \textbf{31.7} & 24.5 & 41.3 & \textbf{39.0} \\
  %\name-\abr{Ensemble} & 37.8 & 29.0 & 32.2  & 25.7 & 31.9 & 41.0 \\
  %Ensemble Calibration (Frequency) & 37.8 & 34.1 & 40.2 & 25.7 & 44.0 & 44.1 \\
  %\name-\abr{Dynamic} &  35.2 & 31.9 & 30.6 & 24.5 & 40.4 & 38.3 \\
  %Consistency Calibration (Frequency) & 35.2 & 44.5 & 35.5 & 24.5 & 46.2 & 40.8 \\
  %Consistency Calibration (Classifier) & 35.2 & 31.6 & 30.6 & 24.5 & 40.3 & 38.7 \\
 \bottomrule
\end{tabular}
\caption{
% \jbgcomment{don't capitalize methods that aren't named after people}
%\sewon{New version}
Results of existing calibration methods
(Section~\ref{subsec:baselines}) and \name\
(Section~\ref{subsec:conscal}). 
% compared to the baselines of logistic
% regression
%Calibration results of baseline calibration methods as well as our new consistency calibration.
% (\abr{lr}), Random Forest (\abr{rf}), and Temperature Scaling
% (\abr{ts}).
%
`\name\ w/o Training Dynamics' is the ensemble-based method
from Section~\ref{subsec:experiments}.  
% We highlight the best result
% in bold.
%Our new consistency calibration method  outperforms all other baselines on \metric{}.
While some existing methods drastically reduce \abr{ece}, none of them significantly reduces \metric; on the other hand, \name\ sets the new state-of-the-art on \metric\ both in-domain and out-of-domain (best results in bold).
Different metrics give different ranking of methods, which further highlights the importance of using a reliable and informative metric.
}
\label{tab:consistency-2}
\end{table*}

%Given the above criticisms against ECE and the conclusion that temperature scaling does not help MacroCE, we examine several other calibration methods to see if they are effective under MacroCE. These methods include: feature-based classifier, neural reranker, and label smoothing. 
% We will show that different metrics give different conclusions about these calibration methods and under the MacroCE metric, none of these methods works well. 
%All experiments are conducted on NQ in-domain data. 

%\sewon{Proposal: right now each subsection has each method + result for that method. What about describing all methods, and then describe all results? This seems to make more sense because our table is also made in this way. Also, for each method, we can only briefly describe and move details into Appendix.}
%\sewon{Another proposal: we could possibly merge Section 5 and 6? Because I think the baselines for the new calibration method in Section 6 should be the methods in Section 5, so it might be good to merge the table and discuss them altogether.}

%\chen{Maybe we can still merge section 5 and 6, as section 5 is relatively short.}
%\sewon{+1}

\subsection{Existing Calibration Baselines}
\label{subsec:baselines}

% \jbgcomment{We're using $p$ for a lot; can we use something else here?}

% \chen{use $t$ instead.}

\paragraph{Simple Baselines.}
We begin with two simple baselines:
\textbf{Binary baseline} assigns the top $t$\% (dev set accuracy) confident predictions in the test set with confidence 1, otherwise 0.
\textbf{Average baseline} assigns all test set predictions with the confidence value equal to the  dev set accuracy.

% \jbgcomment{This seems a little late to bring this up.  Could we hint at this earlier and contrast with the bins?  This can also help motivate the metric we propose.
% But now that I write this, what was the loss function that these trained on.  I'm assuming it's something like L2, which isn't different from our metric.  Let's be clearer about the novelty then (e.g., if it's only for QA, that's fine).
% }
% \chen{What about briefly hint it in section 2?}

\paragraph{Feature Based Classifier.}
%One widely-used calibration method is to train
Prior work has trained a feature-based classifier to predict the
correctness of
outputs~\cite{Zhang2021KnowingMA,Ye2022CanEB}.
%In our experiments, 
%We experiment with feature-based classifiers where we identify a set of potentially relevant features for question answering, and then train a classifier on these features. 
We use SVM to train them as binary classifiers following prior work~\cite{Kamath2020SelectiveQA}. 
% \sewon{Can we say we followed some of prior work on
% this choice?}\chen{If not, what about we only keep one in Table 4
% (maybe put rest of them in appendix)?} \sewon{I agree, in fact, none
% of them reduces ECE.}
%
We include a wide variety of features based on previous
work~\cite{Rodriguez2019QuizbowlTC} (features used are described in
Appendix~\ref{app:details-new-eval}).
During inference time, we use
the classifiers' predicted probability of the test example being
correct as its confidence.

% \jbgcomment{If we have more space, would be good to expand features (just by example).}

\paragraph{Neural Reranker.}
We train a neural reranker as an alternative to manual features.
%Instead of using manual features, an alternative approach is to train a neural answer reranker for post-hoc calibration.
We adopt \abr{ReConsider}~\cite{Iyer2020RECONSIDERRU}
%as our neural answer reranker,
where we train a \abr{bert}-large classifier by feeding in the
concatenation of the question, passage, and answer span.
Passing the raw logit through a sigmoid\footnote{We also experimented
  with softmax but found sigmoid to be substantially better.}
provides the confidence score.
%\sewon{Commented out the softmax and only kept sigmoid.}
%
%We explore two ways of obtaining the confidence value of the top reranked answer prediction: 1) we computer softmax over the top $M=5$ predictions; 2) we compute sigmoid of the raw logit score. 

\paragraph{Label Smoothing.}
In addition to the post-hoc calibration methods above,
%All the above techniques are considered post-hoc calibration because the underlying QA model is fixed and we change the confidence values post-hoc. 
another way of calibration is to train models that are inherently better
calibrated, and a representative approach is label
smoothing~\cite{Pereyra2017RegularizingNN,Desai2020CalibrationOP}.
Label smoothing assigns the gold label probability
$\alpha$ and the other classes $\frac{1 - \alpha}{|Y| - 1}$.
We apply label smoothing on two components of the \abr{odqa}
pipeline: passage selection (where the first passage is gold, the rest
$K - 1$ are false); span selection (where the gold class is the
correct answer start and end position and false classes are the other
positions in the passage).

% \jbgcomment{It looks like only two components are listed here}

\subsection{\name{}: Calibration Through Consistency}
\label{subsec:conscal}

%\chen{It seems that some motivations for consistency calibration is removed, should those be included? }
The failure of temperature scaling under \metric{} implies that only
relying on the final outputs from the \abr{qa} model is not sufficient for
calibration. This calls for additional cues that can reflect the
model's confidence.
We propose to use the model's consistency throughout training as a useful cue.
%
%\jbgcomment{Should also explain the name}
%
We propose a simple calibration method called Consistency Calibration (\name{}).
\name\ compares multiple model checkpoints throughout training and checks if the prediction is consistent.
The intuition is that if the model during training always makes the same prediction, 
%The intuition is that if the same answer prediction is consistent throughout multiple model checkpoints, then it could serve as a strong sign
the model is confident about that prediction, and vice versa.
This is inspired by \textsc{Training Dynamics}
from \citet{Swayamdipta2020DatasetCM};
while they originally use training dynamics
to measure data difficulty, we use it to measure the
confidence of the model predictions.\footnote{Our definition
of consistency is similar to what they call `variability'.}
%While \citet{Swayamdipta2020DatasetCM} aim to measure the difficulty of the examples, we leverage similar intuition but focus on the calibration perspective.\footnote{Note that in \citet{Swayamdipta2020DatasetCM}, confidence is defined as the mean probability of the true label across epochs; our confidence scoring will be unlike their definition.  Moreover, our definition of consistency is similar to they called `variability'.}

Specifically, given $N$ model checkpoints,\footnote{$N=5$ in our
  experiments; we ablate the choice of $N$ in
  Appendix~\ref{app:impact-checkpoint-number}---it has
  marginal impact on the calibration.}  we obtain the final model
prediction $p$ based on the last checkpoint,
% \chenglei{This is true for w/o dynamics but not for the dynamics version tho, for that I just take the final checkpoint' predictions as final predictions. Should we clarify this?}
% \chen{Here we only describe with dynamics I think}
%For both \name{} variants, for each question, we count the majority prediction's occurrence number $n$,  
%we check how many times the predicted answer string also appears from the previous seven checkpoints ($n \in [0,7]$).
%we denote this integer $n \in [0,7]$. 
then count the checkpoints that make the same prediction, and assign a
confidence value 1 if the count is greater than a threshold~$n$;
otherwise 0.
The threshold $n$ is a hyper-parameter chosen
based on the development set.
Apart from this binary confidence
setting, we also explore assigning continuous confidence values based
on checkpoint consistency.
This continuous confidence gets slightly
worse \metric{} than the binary version but improves on all
previous baselines (Appendix~\ref{app:details-new-eval}).

\subsection{Experimental Results}
\label{subsec:experiments}

%\subsection{Results}
%  We use Joint S+E+P as the scoring function with no answer aggregation when deciding the top prediction.
%Apart from the baseline without calibration and the temperature scaling baseline, 
%We also compare with the following other simple baselines to ensure that our claimed improvement on MacroCE is indeed meaningful:
 
% \sewon{Proposal: A bit weird to have these trivial references later. Shouldn't it be introduced when the metric was introduced, or at least when we discuss results in the end of Section 4 or Section 5?}
% \chenglei{How about we combine Table 4 \& 5 so all these are baselines.}
 %\sewon{Another thought 1: I would delete random confidence baseline if we need to save space because it is clearly worse than other two.}
 %\chenglei{Make sense.}
 %\sewon{Another thought 2: Binary and Average baselines are not bad because the dev and the test distributions are similar. What about we add one OOD dataset? I think these methods will have worse OOD result.}
 %\chenglei{We can, although if we are going to add OOD, we might need to add OOD evaluation results for all methods in Table 5. Maybe we can discuss this later in the meeting?}
 %\sewon{Another thought 3: Rename `Average baseline' into `Constant baseline' to make naming consistent to `Binary baseline'? Or maybe both names are not super intuitive.. I'll think more.}
 %\chenglei{Feel free to do naming changes.}
 
Some existing calibration methods lower \abr{ece},
including temperature scaling, the simple average baseline, and label
smoothing (Table~\ref{tab:consistency-2}).
In particular, the simple average baseline has the lowest \abr{ece}
both in-domain and \abr{ood}.
This confirms our earlier point that you can lower \abr{ece} by
assigning confidence values close to the accuracy for all predictions
without any discrimination between correct and wrong predictions.
However, none of these baselines reduces \metric.
%Additional experiment results in Appendix 
%
% Some calibration methods such as
% label smoothing reduces neither of \abr{ece} nor \metric{}, contrary to \citet{Desai2020CalibrationOP}.
%According to Table~\ref{tab:consistency}, similar to temperature scaling, both classfication and neural reranker improve the EM score but they bring no improvement on \metric{}. We conclude that post-hoc approaches do not work well for calibration as it still heavily relies on the model output probabilities.
%In addition, contrary to previous conclusion that label smoothing helps calibration measured by ECE~\cite{Desai2020CalibrationOP}, label smoothing is ineffective for calibration measured by \metric{}.

\jbgcomment{I really dislike the verb ``outperform''.  Say the metric
  you're using, particularly when there are multiple in this paragraph}
\name{} significantly lowers \metric{}, outperforming the previous
best by 11\% and 6\% absolute in-domain and out-of-domain,
respectively.
This confirms the effectiveness of using consistency over different
checkpoints throughout training.
Notably, that \name{} significantly outperforms the simple binary
baseline highlights that binary confidence is not the reason for
better \metric{}.
Additional results in
Appendix~\ref{app:more_baselines} compare the joint and pipeline
approach (defined in section~\ref{subsec:base-calibration}), which
have similar \metric{}.

\begin{figure}[t]
 \centering
 \includegraphics[trim=0.1cm 0.1cm 0.1cm 0.1cm,clip=true,width=1.01\linewidth]{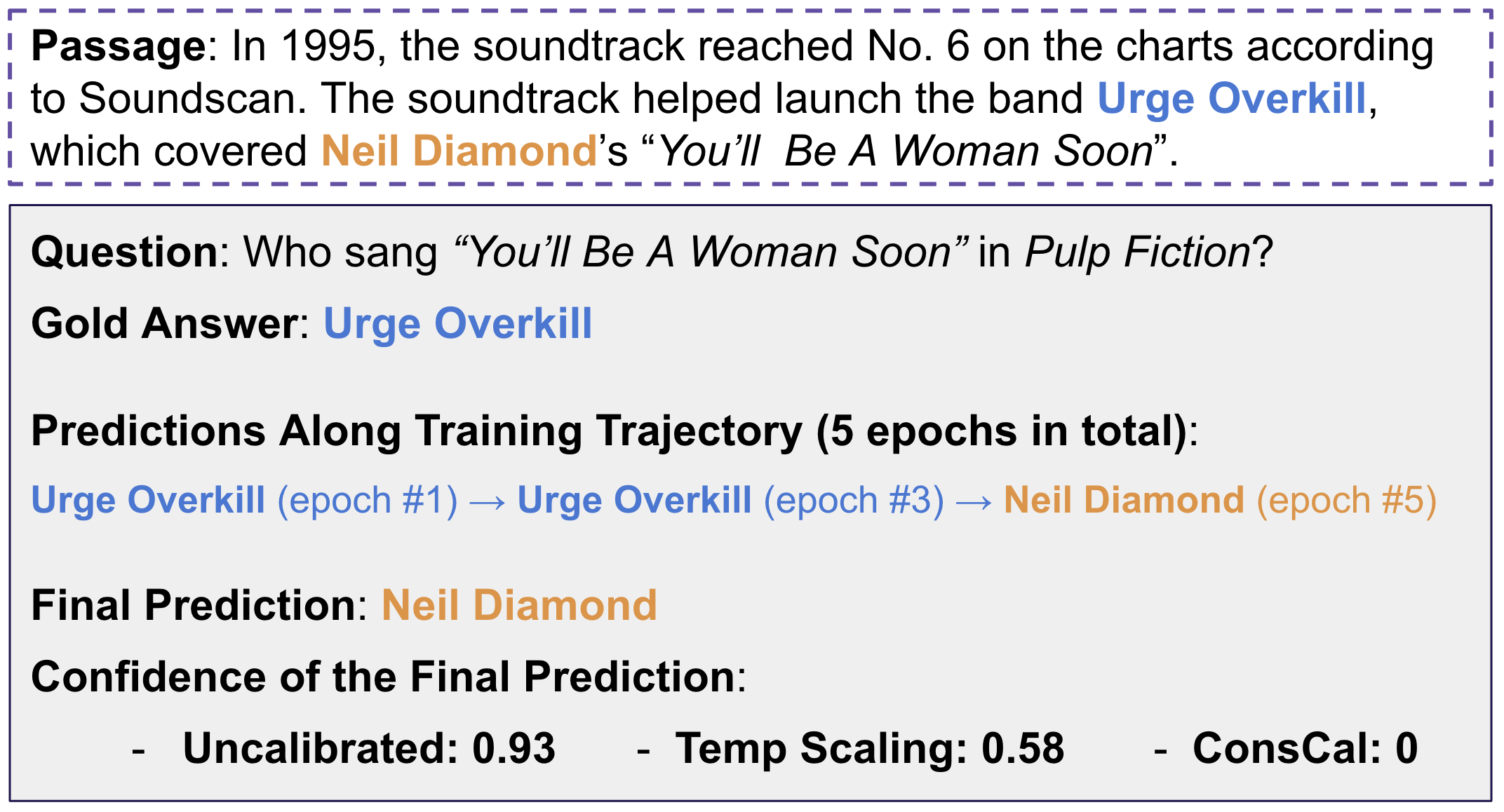}
\caption{An example from \abr{nq} where the final prediction is wrong.
%The intermediate checkpoints predicted ``Urge Overkill'' while the final checkpoints predicted ``Neil Diamond'', which is wrong.
The original uncalibrated confidence is over-confident, while
temperature scaling lowers it, it is still over-confident
and could mislead users.
% \name{} calibrates the confidence to 0 based on the inconsistency across checkpoints to reveal that the model is not confident about the final prediction.
\name{} sets the confidence to 0 because predictions across checkpoints in the training trajectory are inconsistent---the model is confused between the specific performer and the original singer in this example.
%based on the inconsistency across checkpoints to reveal that the model is not confident about the final prediction.
% \jbgcomment{This example would be stronger with: A) a plot of the training trajectory B) it would be better still if that showed the prediciton changing C) explicitly explaining why the model is confused: model is confused between specific performer and original singer.} 
% \jbgcomment{Maybe also have another example with how it can go wrong?  And include other calibration metrics?}
}
\label{fig:example}
 \end{figure}

To analyse whether the gains of \name{} are simply from ensembling multiple
checkpoints, we compare with an additional baseline called \name\
w/o Training Dynamics: we finetune the model $N$ times
independently using different random seeds,\footnote{Thus, this
baseline has $N$ times larger training cost.} obtain final predictions through majority vote, and
compute the confidence scores as we did for \name.
We use the same $N$
($N=5$) for both \name{} and \name\ w/o Training Dynamics. % so that
% it is a fair comparison.
%
While this variant reduces \metric\
more than any of previous methods, its \metric\ values are still higher
than \name\ both in-domain and \abr{ood}.
This suggests that,
while ensembling is one factor for reducing \metric, it
is not the only factor---considering training dynamics remains important.
We provide a qualitative example in Figure~\ref{fig:example}: the
model changes its prediction in the last epochs -- a cue to suggest low confidence via \name{}.

% \sewon{So are we using $N=5$ for ensemble method, while we use $N=16$ for training dynamics?} \chenglei{Changing that to use N=5 for both, I will add explanation in a minute.}

%Compared to existing approaches, both \name{}-\abr{Ensemble} and \name{}-\abr{Dynamic} significantly lower \metric{}, which confirms that consistency provides a stronger signal towards better calibration. Furthermore, we observe that \name{}-\abr{Dynamic} has lower \metric{} over \name{}-\abr{Ensemble}, showing that training dynamics is more informative than multiple final model checkpoints.

%achieves the best MacroCE compared to all other baseline calibration methods. We believe that it is possible to develop more sophisticated calibration methods based on our consistency idea to further improve the calibration performance, and we leave it to future work. 

% %
% \jbgcomment{We really need some qualitative examples.  E.g., have an example, go through the features in LR etc. and how they don't show that it's uncertain, but show the trajectory during training and show how that screws everything up.}
% \chenglei{Not sure if we have space for this and what's the best way of presenting this. What do others think?}

% \chen{It maybe good to show an instance with consistency Calibration, but I agree space is an issue}

\begin{table*}[t]
\small
\begin{center}
\setlength{\tabcolsep}{2.5mm}{
\begin{tabular}{ l c c | c c c c}
 \toprule
  & ECE & \metric & Precision & Recall & F1 & Agreement ($\alpha$) \\
 \midrule
No Confidence  & -- & -- & 0.37 & 0.62 & 0.46  & 0.21 \\
Raw Confidence  & 30.4 & 44.5 & 0.38 & 0.67 & 0.48 & 0.35 \\
Temp Scaling & 4.7 & 42.5 & 0.44 & 0.72 & 0.54 & 0.33 \\
\name  & 33.1 & 31.7 & 0.50 & 0.68 & 0.58 & 0.40 \\
\name~ (Always Trust) & 33.1 & 31.7 & 0.53 & 0.82 & 0.64 & -- \\
 \bottomrule
\end{tabular}}
\caption{We ask humans---given an estimate of confidence---whether
  they think a \abr{qa} system is correct or not.  The model
  accuracy on this sampled set is 35\%.  Apart from human ratings, we
  additionally show a baseline of always following \name's judgement
  in the last row.  Showing the confidence significant improves human
  judgement, especially with \name. Surprisingly, annotators sometimes
  do not trust and overrule \name's confidence scores, which results in worse F1 (second to last row).
  Furthermore, \metric~ ranks \name~ as the best method, agreeing with
  human evaluation, while \abr{ece} misleadingly favors temperature scaling.
}
 \label{tab:human_study}
\end{center}
\end{table*}

 \subsection{Human Study}

Finally, we investigate whether \name{} improves user decision
making---mimicing validating a search engine's answer to a
question---with a human study.
We randomly sample 100 questions from the \abr{nq} test set and
present the questions to annotators along with the \abr{dpr-bert}
predictions.
We ask annotators to judge the correctness of the model predictions
under four settings:
(1) show only questions and predictions without model confidence;
(2) show the \abr{qa} pairs along with raw model confidence without
calibration;
(3) show the \abr{qa} pairs along with temperature scaled confidence;
(4) show the \abr{qa} pairs along with confidence calibrated by
\name{}.
We recruit a total of twenty annotators (five annotators under each
setting) on Prolific, each annotating 100 questions, with average
compensation of \$14.4/hour.

We measure the precision, recall, and F1 score of the human judgement.
We also report Krippendorff's alpha among the five
annotators\footnote{The Krippendorff's alpha indicates only moderate
  agreement, which is expected because different people have different
  preferences when to trust an answer. Nonetheless,  \name{} increases
  the agreement among human annotators.}.
Showing the confidence scores significantly improves human decision
making (Table~\ref{tab:human_study}), and \name{} helps achieve better
F1 than temperature scaling.
Interestingly, despite \name{} providing binary confidence scores of 0
and 1, humans sometimes do not follow the confidence scores and
``overrule'' them. This actually leads to worse F1 than a baseline
that always follows \name{}'s confidence.
Moreover, the \abr{ece} metric ranks temperature scaling as best, contradicting human study results.  Nevertheless, our
\metric{} correctly ranks \name{} as the best method, aligning with
human judgement.

% \section{Beyond Question Answering}
% \chenglei{TODO: Add results for SST and NLI.}

\section{Related Work}\label{sec:related}

% \jbgcomment{Did not take a pass on this, section obviously not complete}
% \sewon{Missing discussion: other calibration metrics proposed in previous work? Other than ECE.}
% \chen{I will leave it for Chenglei to add some contents first}
% %\paragraph{Model Calibration.}

%\chen{Let's be consistent and use present tense for this section?}

% \jbgcomment{I think this could use a bit of a discussion up front about what's included, how it's organized, etc.}

This section reviews prior work on calibration metrics and methods that are relevant to \abr{nlp}. 

\paragraph{Calibration Metrics.} Brier Score~\cite{Brier1950VERIFICATIONOF} is one of the earliest calibration metrics that sums over squared errors between accuracy and confidence for all instances, but is only applicable to binary classification. \citet{Naeini2015ObtainingWC} first use bucketing mechanisms to compute calibration errors, and propose metrics like \abr{ece} and maximum calibration error (\abr{mce}), \citet{Nguyen2015PosteriorCA} explore other bucketing mechanisms like equal-mass binning. 
\citet{nixon2019measuring} point out the flaws of the fixed range bucketing mechanism  used by \abr{ece}, and propose an adaptive bucketing mechanism as an alternative.
We take inspiration from these prior analysis, identify the problems grounded in a more complex open-domain question answering task, and propose \metric{} to address the issues.
%\citet{minderer2021revisiting} note the discrepancy in results when using different calibration metrics. 
%Our proposed \metric{} metric takes inspiration from these prior analysis and addresses the pitfalls of bucketing based metrics like \abr{ece}. 

\paragraph{Calibration Methods.}
%
% Model calibration has been studied in statistical machine learning. 
\citet{guo2017calibration} first experiment various post-hoc calibration methods including temperature scaling, and
\citet{Thulasidasan2019OnMT} find that mixup training improves the calibration of image classifier evaluated by the \abr{ece} metric. 
%
% \citet{nixon2019measuring} focused on image classification and claimed that different calibration metrics influence calibration methods. 
% %\citet{minderer2021revisiting} benchmarked calibration of more recent vision models. 
% \citet{Thulasidasan2019OnMT} find that for image classification, using mixup training improves calibration evaluated by the ECE metric. 
%
%\paragraph{Calibration in NLP.}
Within \abr{nlp} domain, previous works have explored calibration on various tasks, including multi-class classification~\cite{Desai2020CalibrationOP} and sequence tagging~\cite{Nguyen2015PosteriorCA}. In question answering, \citet{Kamath2020SelectiveQA} propose the selective question answering setting that aims to abstain as few questions as possible while maintaining high accuracy.
%
%measure the \abr{qa} model accuracy while abstaining the least confident questions. 
Toward this goal, 
later approaches~\cite{Zhang2021KnowingMA,Ye2022CanEB} use featured based approach to train a binary classifier for deciding what questions to abstain. 
%In a similar vein, \citet{Rodriguez2019QuizbowlTC} explore optimal buzzing strategy in the setting of incremental QA (Quizbowl). 
%
While selective question answering offers a way of measuring calibration, the scale of confidence values is not considered since abstention can be effective as long as correct predictions have higher confidence than wrong ones, regardless of the absolute scales. 
%
% In this work, we focus on MacroCE instead since it consider
% An important distinction between them and our work is that selective question answering only cares about the relative ranking of the answer predictions rather than 
% the absolute calibration error. 
% \chen{I do not understand is sentence}
%
%
Concurrent work~\cite{Dhuliawala2022CalibrationOM}  explores
calibration for retriever-reader \abr{odqa}, focusing on combining
information from the retriever and reader.  In addition to
span-extraction \abr{qa}, \citet{jiang2021can} and \citet{si2022prompting} also explore
calibration for generative \abr{qa}. 
However, these works use \abr{ece} for evaluation. 
% While we expect that our consistency calibration method can also be adapted on generation settings, we leave this for future exploration. Is is worth noting that all these works used bucketing-based ECE metric for evaluation. 

% The most important distinction between our work and the above is that we use our new MacroCE metric as the primary evaluation metric. This reveals that some of the methods previously shown to be effective are actually not useful. 

% \chen{Did a quick edit, I think this section is a bit too scattered, we introduced related work here and there, but most are separated.}

\section{Conclusion}
%\jbgcomment{The conclusion could be much stronger.  Some points you could hit:
%\begin{itemize}
%        \item Calibration is generally a problem for explainable AI
%        \item Current metrics do not emphasize the important decision that a user needs to make when they see an output: do I trust this or not
%        \item We propose a general metric that gets to that question
%        \item We investigate this problem in the very real situation of trusting the output of a QA system (you could cite Shi's augment paper for evidence that just showing probability isn't that helpful, and that might be a good piece of future work to point to how to actually evaluate this)
%        \item I don't know if the conclusion is the right place for this, but somewhere in the paper you should go through some of the examples of crazy QA that has come out of Google Assist, Alexa, Siri, etc. to highlight the problem of whether they are trustworthy.
%        \item Then talk about how this improves QA (going back to the qualitative example)
%        \item We have some ideas about generation, would be good to talk about that in vague terms here.
%\end{itemize}
%}

% Calibration is one of the core challenges towards  trustworthy and explainable  AI. 
This paper investigates calibration in the realistic application of \abr{odqa} where users need to decide whether to trust the model prediction based on the confidence scores. 
Although confidence scores produced by existing calibration methods improves the popular \abr{ece} metric,  
%We then challenge the conventional view of calibration and the use of ECE as the metric. 
these confidence scores do not help distinguish correct and wrong predictions. 
We propose to use the \metric{} metric to remedy the flaws, 
% Our controlled experiments demonstrate the advantage of \metric{} over \abr{ece}, and 
and existing calibration methods fail on our \metric{} metric.
%
%Towards better calibration,
We further propose a simple and effective calibration method \name{} that leverages training consistency. 
Our human study confirms both the effectiveness of \name{} as well as the alignment between \metric{} and human preference. 
Our work advocates and paves the path for user-centric calibration, and our \name{} method is a promising direction for better calibration.
Future work can adapt our calibration metric and method to more
diverse tasks (such as generative tasks) and explore other ways to
further improve user-centric calibration.

%\jbgcomment{Future work: human evaluation, other tasks, focusing on different parts of the pipeline.}

\section*{Limitations}\label{sec:limitation}
% \chen{I suggest talking about the following points, feel free to add more: 
% 1) about metric, our metric does not work for every use case, it cannot handle cases where we want to have 0.5 confidence, e.g., situations where labels are ambiguous
% 2) our calibration evaluation is limited to QA generation task, talk about their eval metric? 
% }

% \jbgcomment{More citations would make this stronger; it's free space!}
We note several limitations of this paper and point to potential future directions to address them:

% \sewon{Reduce the words a bit.}
\begin{itemize}
    \item %Our proposed
    \metric{} is motivated from a user-centric perspective where we want to maximally distinguish correct and wrong predictions.  However, it is not the panacea for all use cases, e.g., in some applications, the %ideal 
    confidence output might have to be mediocre %be around $0.5$
    to indicate the uncertainty of the output, rather than taking a stance as \metric\ encourages.
    %which is not encouraged by \metric{}. 
    %Future work may also consider conducting human studies to confirm what other use cases make \metric{} more favorable than previous metrics.  
    
    \item Our experiments are focused on \abr{odqa} and in particular, %we use
    span-extraction models (we also showed similar findings on binary sentiment classification in Appendix~\ref{app:more_ece}). 
    While we expect %the issues about \abr{ece}
    most findings in this paper
    to hold for other models and tasks as well, this needs to be empirically verified in future work.
    In particular, one promising line for future work is to verify whether %our
    \name{} also works well for text generation tasks and models. 
\end{itemize}

\section*{Ethical Considerations}\label{sec:ethics}
\paragraph{Data and Human Subjects} All datasets used in this paper are from existing public sources and we do not expect any violation of intellectual property or privacy. All human annotators that we recruited on Prolific are well-compensated and we did not receive any complaints from the annotators regarding the job (we do not perceive any possible harm on them either).  

\paragraph{Broader Impact} 
We expect our study to have a positive impact on the safe deployment of AI applications.
Our study is targeted towards the real-world application of question answering from a user-centric perspective. 
We make model predictions more trustworthy to users by providing well-calibrated confidence scores. 
This especially helps users avoid misleading wrong predictions which can cause serious troubles in real-life applications such as digital assistants and search engines. 
Our human study has also confirmed the advantages of our proposed metric and calibration method. 
% We believe that our proposed calibration metric better captures user needs and our proposed method better satisfies user needs. 
% Taken together, we foresee our work to bring positive impact to language technology users especially in high-stake settings when users need to decide which model predictions to trust and base their decisions on. 
% \sewon{I think this particular paragraph is not super informative because it repeats what we've said in the intro, etc. Maybe we could just cut the paragraph or shorten it a lot?}

\section*{Acknowledgement}
We thank Yanai Elazar, Haozhe An, Yichong Xu, He He and Kyunghyun Cho for their helpful discussion and feedback.
This work is supported by NSF
Grant IIS-1822494. Any opinions, findings, conclusions, or recommendations expressed here are
those of the authors and do not necessarily reflect
the view of the sponsors.

\bibliography{bib/journal-full, bib/custom,bib/jbg}
\bibliographystyle{acl_natbib}

\appendix
\clearpage

\section*{Appendix}

\section{More Analysis on Calibration Metrics}

\paragraph{Temperature Scaling with Different Temperature Scalars}

% \sewon{I think I would cut this paragraph if we are over 8 pages.}
% \chen{Agree, let's leave it here first.}

% \chenglei{Will try to make plots look nicer later. It should be clear even without colors. Also: unify the naming of different metrics.}

\begin{figure}[t]
    \centering
    \includegraphics[width=0.5\textwidth]{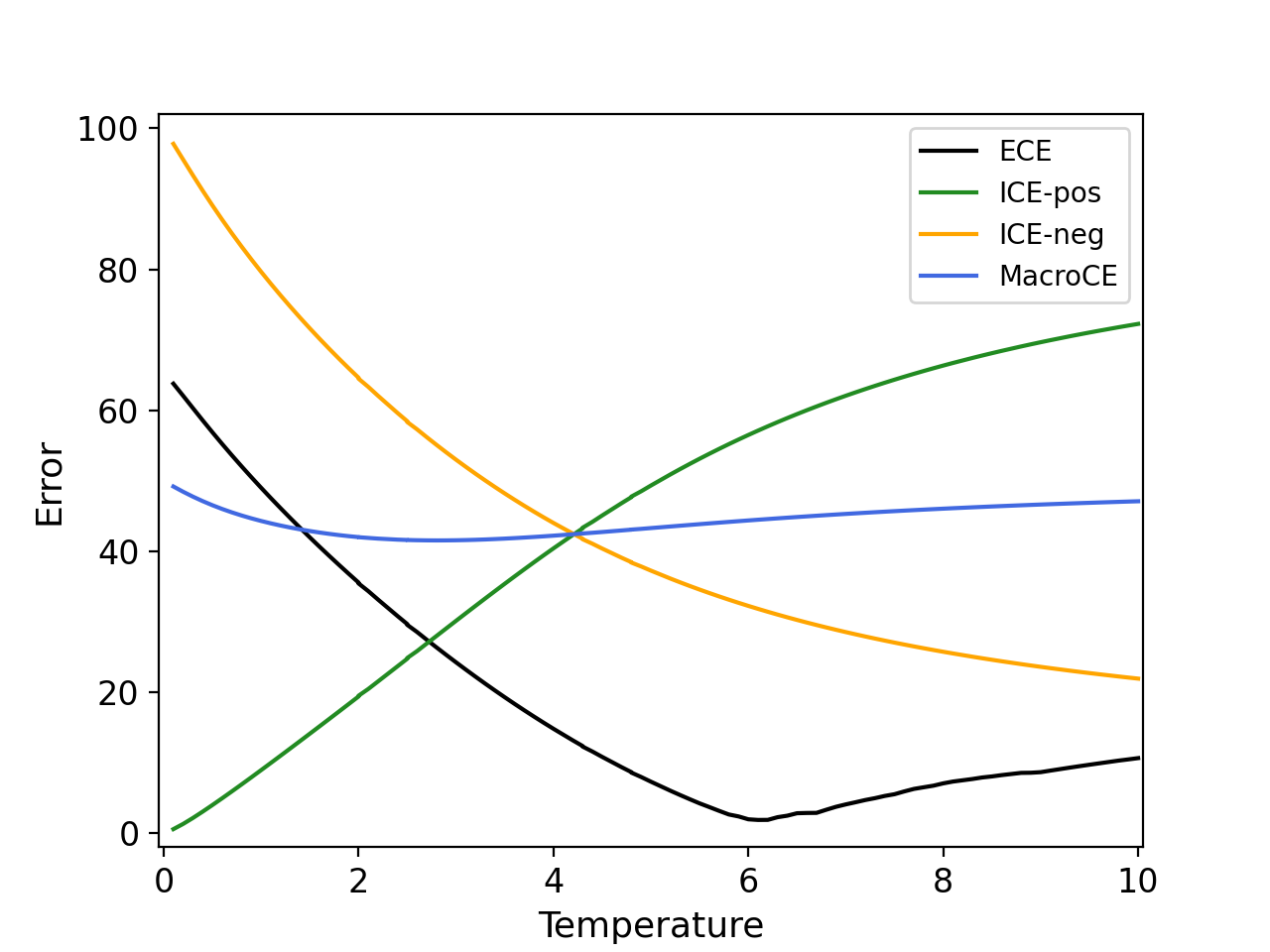}
    \caption{Calibration errors after temperature scaling. x-axis represents different temperature values; lines with different colors represent different metrics. \metric{} stays relatively constant while \abr{ece} varies largely at different temperature values.}
    \label{fig:all_temperatures}
\end{figure}

%vary the temperature scalar $\tau$ from 0.01 to 10.0 with a 0.01 interval and 
We apply temperature scaling with varying temperature scalar $\tau$.
% This offers a direct comparison between these metrics and also allow us to understand the impact of temperature scaling. 
% The model accuracy (EM) stays at 34.18. %
According to Figure~\ref{fig:all_temperatures},
as we increase the temperature value, the confidence scores decrease, and consequently \abr{ice}$_{\mathrm{pos}}$ increases and \abr{ice}$_{\mathrm{neg}}$ decreases. Meanwhile, \metric{} stays constant while \abr{ece} changes drastically, which reflects the flaw of temperature scaling: a single temperature value cannot improve calibration for both correct and wrong predictions simultaneously. Such flaw is only captured by \metric.

\paragraph{ECE with Equal-Width and Equal-Mass Binning}
We compare measuring calibration with equal-width bucketed ECE and equal-mass bucketed ECE in Table~\ref{tab:temp_scale_additional}. We find that both variants of ECE give similar results on all experiment settings, and both of them  show contrary conclusions than \metric{} (\textit{e.g.}, they both underestimate the calibration errors of temperature scaling in OOD settings).

\paragraph{Calibration Results at Different Accuracy } In Table~\ref{tab:analysis2_app}, we show the numerical results of \abr{ece}, \abr{ice}, and \metric{} at different accuracy. The results show that only \metric{} is stable. 

\paragraph{Calibration Results under Accuracy Shift} In Table~\ref{tab:analysis3_app}, we show numerical calibration results under accuracy shifts where the dev and test accuracy differ largely. \metric{} is the only metric that stays stable under such shifts. 

\section{Implementation Details of Methods in Section~\ref{sec:new-eval}}\label{app:details-new-eval}

\paragraph{Feature Based Classifier}
We include the following features based on previous work~\cite{Rodriguez2019QuizbowlTC}: 
the length of the question, passage and predicted answer; raw and softmax logits of passage, span position selection; softmax logits of other top predicted answer candidates; the number of times that the predicted answer appears in the passage and the question; the number of times the predicted answer appears in the top candidates.

We use the QA model's predictions on the NQ dev set as the training data. We re-sample the data to get a balanced training set, and the training objective is binary classification on whether the answer prediction is correct. 
We hold out 10\% predictions as the validation set and we apply early stopping based on the validation loss. 
During inference, we directly use the predicted probability as the confidence value.

\paragraph{Neural Reranker}
During training, for each question we include one randomly chosen positive and $M - 1 (M=10)$ randomly chosen hard negatives (hard negatives are negative predictions with the highest raw logits). We use \abr{dpr-bert}'s predictions on the \abr{nq} training set for reranker training. During inference, we use the trained reranker to rerank the top five predictions. In particular, we use sigmoid to convert the raw reranker logits to probabilistic confidence values.

\paragraph{Label smoothing}
We use $\alpha = 0.1$ in our experiments, and we find that the calibration results are largely insensitive to the choice of $\alpha$. 
We change the loss function from cross entropy to KL divergence with the label smoothed gold probability distribution. 
We compare the calibration results of the model trained without and with label smoothing, and we also explore applying temperature scaling on top of the model trained with label smoothing. 

\paragraph{\name{}}
We use the final checkpoint's predictions as the final predictions. 
We do this instead of taking the majority vote of all intermediate checkpoints before earlier checkpoints have lower answer accuracy than the final checkpoint.
% compare two ways of obtaining the top prediction of each checkpoint - joint (considering top predictions from all top-10 retrieved passages) and pipeline (only considering the top predictions from top-1 retrieved passage). 

\begin{table}[t]
\small
\begin{center}
\setlength{\tabcolsep}{1.3mm}{
\begin{tabular}{ l c @{\hspace{3\tabcolsep}} cccc }
    %& \multicolumn{2}{c}{$\overbrace{\phantom{~~~~~}}^{\text{Section~\ref{sec:exp}}}$} & \multicolumn{2}{c}{$\overbrace{\phantom{~~~~~~~~}}^{\text{Section~\ref{sec:exp-new-metrics}}}$} \\
    % && \multicolumn{3}{c}{Section~\ref{sec:calibrate_qa}} & \multicolumn{2}{c}{Section~\ref{sec:exp-new-metrics}} \\
 \toprule
    Model & TS & EM$_{\uparrow}$ & ECE$_{\textrm{width}\downarrow}$ & ECE$_{\textrm{mass}\downarrow}$ & \metric{}$_{\downarrow}$ \\ 
 \midrule
    \multicolumn{5}{l}{\textbf{\em NQ}} \\
    Joint & -  & 32.9 & 27.1 & 27.1 &  43.6 \\
    Joint & \checkmark & 32.9 & 4.0 & 4.3 &  \textbf{42.5} \\
    Pipeline & - & 34.1 & 48.2 & 48.2 &  44.2 \\
    Pipeline & \checkmark & 34.1 & \textbf{2.7} & \textbf{3.2} &  44.4 \\
 \midrule
    \multicolumn{5}{l}{\textbf{\em NQ $\rightarrow$ \abr{HotpotQA}}} \\
    Joint & - & 24.9 & 41.0 & 41.0 &  45.7 \\
    Joint & \checkmark & 24.9 & 12.5 & 12.4 &  \textbf{45.5} \\
    Pipeline & - & 22.6 & 59.6 & 59.6 & 47.4 \\
    Pipeline & \checkmark & 22.6 & \textbf{8.4} & \textbf{8.4} & 47.7 \\ 
 \midrule
    \multicolumn{5}{l}{\textbf{\em NQ $\rightarrow$ \abr{TriviaQA}}} \\
    Joint & - & 33.6 & 25.4 & 25.4 &  45.1 \\
    Joint & \checkmark & 33.6 & 6.4 & 6.5 &  44.3 \\
    Pipeline & - & 34.2 & 48.2 & 48.2 &  \textbf{43.7} \\
    Pipeline & \checkmark & 34.2 & \textbf{6.1} & \textbf{5.9} &  44.6 \\
    \midrule
    \multicolumn{5}{l}{\textbf{\em NQ $\rightarrow$ \abr{SQuAD}}} \\
    Joint & - & 12.4 & 41.7 & 41.7 &  \textbf{39.5} \\
    Joint & \checkmark & 12.4 & \textbf{12.4} & \textbf{12.4}  &  39.7 \\
    Pipeline & - & 12.2 & 62.7 & 62.7 &  41.4 \\
    Pipeline & \checkmark & 12.2 & 13.5 & 13.5 &  43.9 \\
 \bottomrule
\end{tabular}}
 \caption{
    We compare ECE with equal-width binning (ECE$_{\textrm{width}}$) and equal-mass binning (ECE$_{\textrm{mass}}$) on both in-domain and \abr{ood} evaluation. They share the same pattern on all cases. 
 }
 \label{tab:temp_scale_additional}
\end{center}
\end{table}

\begin{table}[t]
\small
\begin{center}
\setlength{\tabcolsep}{1.5mm}{
\begin{tabular}{ l c c c c }
 \toprule
  & Temp & ECE & ICE & MacroCE \\ 
 \midrule
    \multicolumn{5}{l}{\textbf{\em Acc=10\%}} \\
    Before Calibration & 1.00 & 69.97 & 71.93 &  44.32 \\
    After Temp Scaling & 10.00 & 11.49 & 26.08 & 46.90 \\
 \midrule
    \multicolumn{5}{l}{\textbf{\em Acc=50\%}} \\
    Before Calibration & 1.00 & 34.15 & 44.50 & 44.50 \\
    After Temp Scaling & 4.27 & 5.60 & 42.42 & 42.42 \\
 \midrule
    \multicolumn{5}{l}{\textbf{\em Acc=90\%}} \\
    Before Calibration & 1.00 & 7.96 & 8.08 & 45.33 \\
    After Temp Scaling & 0.47 & 9.08 & 13.18 & 47.77 \\
 \bottomrule
\end{tabular}}
 \caption{Calibration results where we re-sample the predictions to vary the model accuracy (10\%, 50\%, 90\%; same on both development and test sets). ``Temp'' represents the temperature value tuned on the development set. \abr{ece} results before calibration vary at different accuracy.}
 \label{tab:analysis2_app}
\end{center}
\end{table}

\begin{table}[t]
\small
\begin{center}
\setlength{\tabcolsep}{2mm}{
\begin{tabular}{ l c c c c }
 \toprule
  & Temp & ECE & ICE & \metric \\ 
 \midrule
    \multicolumn{5}{l}{\textbf{\em Development=90\% $\rightarrow$ Test=10\%}} \\
    Development & 0.47 & 7.20 & 12.29 & 46.64 \\
    Test & 0.47 & 80.54 & 81.49 & 47.39 \\
 \midrule
    \multicolumn{5}{l}{\textbf{\em Development=10\% $\rightarrow$ Test=90\%}} \\
    Development & 10.00 & 12.23 & 26.88 & 47.48 \\
    Test & 10.00 & 63.87 & 68.00 & 46.94 \\
 \bottomrule
\end{tabular}}
 \caption{Calibration results when training and test accuracy are different. In the first case, we tune the temperature value on a development set with only 10\% correct predictions and a test set with 90\% correct predictions, and we reverse the setup in the second case. 
\abr{ece} and \abr{ice} change significantly under such accuracy shifts even though the underlying model is the same. In contrast, only \metric{} is stable under train-test accuracy shifts as desired.}
 \label{tab:analysis3_app}
\end{center}
\end{table}

\begin{table*}[t]
\small
\setlength\tabcolsep{5pt}
\centering
\begin{tabular}{ c c c c c c | c c c c c} 
& \multicolumn{5}{c}{IID (NQ)} & \multicolumn{5}{c}{OOD (HotpotQA)} \\
 \toprule
 Calibrator &  
 EM & ECE\textsubscript{interval} & ICE\textsubscript{pos} & ICE\textsubscript{neg} & MacroCE & EM & ECE\textsubscript{interval} & ICE\textsubscript{pos} & ICE\textsubscript{neg} & MacroCE
 \\
  \midrule
  \multicolumn{8}{l}{\textbf{\em No Calibration}} \\
   Joint & 35.24 & 30.41 & 27.17 & 61.74 & 44.46 & 24.50 & 44.40 & 25.38 & 67.04 & 46.21 \\
    Pipeline & 36.34 & 50.08 & 6.93 & 82.63 & 44.78  & 22.78 & 61.22 & 12.48 & 82.95 & 47.72 \\
    \midrule
\multicolumn{8}{l}{\textbf{\em Binary Baseline}} \\
    Joint & 35.24 & 38.03 & 53.22 & 29.77 & 41.50 & 24.50 & 38.62 & 55.81 & 33.05 & 44.43 \\
    Pipeline & 36.34 & 33.68 & 46.34 & 26.46 & 36.40  & 22.78 & 38.82 & 55.45 & 33.92 & 44.68 \\
\midrule
\multicolumn{8}{l}{\textbf{\em Average Baseline}} \\
    Joint & 35.24 & 2.00 & 64.25 & 35.75 & 50.00 & 24.50 & 11.25 & 64.25 & 35.75 & 50.00 \\
    Pipeline & 36.34 & 0.03 & 63.66 & 36.34 & 50.00 & 22.78 & 13.56 & 63.66 & 36.34 & 50.00 \\
\midrule
\multicolumn{8}{l}{\textbf{\em Temperature Scaling}} \\
Joint &  35.24 & 4.69 & 53.98 & 31.08 & 42.53 &  24.50 & 13.72 & 55.17 & 36.07 & 45.62  \\
 Pipeline & 36.34 & 5.49 & 58.57  & 30.54  & 44.55 & 22.78 & 8.14 & 65.78 & 29.95 & 47.86 \\
 \midrule
\multicolumn{8}{l}{\textbf{\em Feature-based Classifier}} \\
Logistic Regression & 37.26 & 61.50 & 0.04 & 98.04 & 49.04 & 21.37 & 76.56 & 0.46 & 97.39 & 48.93 \\
SVM & 36.51 & 52.30 & 4.88 & 85.18 & 45.03 & 21.82 & 62.40 & 10.93 & 82.87 & 46.90 \\
Random Forest & 36.40 & 47.41 & 9.57 & 80.02 & 44.79 & 24.61 & 53.88 & 16.78 & 76.95 & 46.87 \\
  \midrule
\multicolumn{8}{l}{\textbf{\em Neural Reranker (ReConsider)}} \\
Softmax  & 37.62  & 33.83 & 33.70 & 62.24 & 50.47 & 26.54 & 46.80 & 27.84 & 69.79 & 48.81 \\
Sigmoid & 37.62 & 58.61 & 9.37 & 72.59 & 40.97 & 26.54 & 51.41 & 23.56 & 70.40 & 46.98 \\
 \midrule
\multicolumn{8}{l}{\textbf{\em Label Smoothing}} \\
Joint & 36.12 & 29.42 & 28.81 & 62.36 & 45.59 & 23.57 & 44.69 & 25.88 & 67.76 & 46.82 \\
 + TS  & 36.12 & 5.57 & 56.41 & 30.48 & 43.45 & 23.57 & 14.31 & 56.27 & 35.82 & 46.04 \\
  \midrule
\multicolumn{8}{l}{\textbf{\em Ensemble Calibration (Binary)}} \\
Joint  & 37.78 & 28.95 & 45.23 & 19.06 & 32.15  & 25.68 & 31.86 & 59.93 & 22.15 & 41.04 \\
 Pipeline & 38.98 & 29.22 & 45.63 & 18.75 & 32.19 & 23.43 & 32.17 & 58.48 & 24.11 & 41.30 \\
   \midrule
\multicolumn{8}{l}{\textbf{\em Ensemble Calibration (Frequency)}} \\
Joint  & 37.78 & 34.12 & 15.95 & 64.53 & 40.24 & 25.68 & 43.95 & 21.62 & 66.62 & 44.12 \\
 Pipeline & 38.98 & 33.00 & 15.93 & 64.26 & 40.10 & 23.43 & 47.03 & 21.16 & 67.89 & 44.52 \\
\midrule
\multicolumn{8}{l}{\textbf{\em Consistency Calibration (Binary)}} \\
Joint &  35.24 & 31.91 & 26.10 & 35.07 & 30.59 & 24.50 & 40.37 & 34.29 & 42.35 & 38.32 \\
Pipeline & 36.34 & 31.30 & 25.91 & 34.38 & 30.15 & 22.78 & 43.25 & 37.00 & 45.09 & 41.05 \\
\midrule
\multicolumn{8}{l}{\textbf{\em Consistency Calibration (Frequency)}} \\
Joint & 35.24 & 44.46 & 26.39 & 44.68 & 35.53 & 24.50 & 46.21 & 31.72 & 49.89 & 40.81 \\
Pipeline & 36.34 & 44.78 & 26.04 & 43.78 & 34.91 & 22.78 & 47.72 & 32.34 & 51.87 & 42.11 \\
\midrule
\multicolumn{8}{l}{\textbf{\em Consistency Calibration (Classifier)}} \\
Joint & 35.24 & 31.63 & 27.28 & 34.00 & 30.64 & 24.50 & 40.34 & 35.56 & 41.90 & 38.73 \\
Pipeline & 36.34 & 33.16 & 25.84 & 37.34 & 31.59 & 22.78 & 44.60 & 31.81 & 48.38 & 40.09 \\
 \bottomrule
\end{tabular}
\caption{Calibration results of baseline calibration methods as well as our new consistency calibration. We compare using the joint and pipeline approach for obtaining the top answer predictions. We highlight the best result in bold. Our new consistency calibration method  outperforms all other baselines on \metric{} for both the joint and pipeline implementation. Note that different metrics give different ranking of these methods, which further highlights the importance of using a reliable and informative metric. }
\label{tab:consistency}
\end{table*}

\section{Dataset Details}
\label{sec:datasets}

\noindent
\textbf{Natural Questions (NQ)}~\citep{kwiatkowski-19} consists of questions mined from Google search queries. We use the open version of NQ where each question has answers with up to five tokens found from Wikipedia~\citep{lee2019latent}. We use NQ for training and in-distribution evaluation.

\noindent
\textbf{SQuAD}~\citep{rajpurkar-16} contains a set of questions written by crowdworkers given a Wikipedia paragraph. We use the open version of SQuAD following \citet{chen2017reading}. We use SQuAD for out-of-distribution evaluation.

\noindent
\textbf{TriviaQA}~\citep{joshi-17} includes trivia questions scraped from the web. We use the unfiltered version for out-of-distribution evaluation.

\noindent
\textbf{HotpotQA}~\citep{yang2018hotpotqa} is a multi-hop question answering dataset written by crowdworkers given a pair of Wikipedia paragraphs. We take the full-wiki version of HotpotQA and use it for out-of-distribution evaluation.

\section{Additional Baselines and Variants}
\label{app:more_baselines}

We compare two ways of obtaining the top prediction of each checkpoint - joint (considering top predictions from all top-10 retrieved passages) and pipeline (only considering the top predictions from top-1 retrieved passage). In the main paper (Table~\ref{tab:consistency-2}) we reported results of using the joint approach. We compare these two variants in Table~\ref{tab:consistency}. 

We also explore two other alternative implementations of \name{}. The first one is frequency-based \name{} where the confidence of each prediction is computed as $\frac{k}{n}$ where $k$ is the number of times that the prediction was made by the $n$ total checkpoints. The second one is classifier-based \name{} where we treat whether the final prediction is also predicted by each intermediate checkpoint as a binary feature, and train a linear classifier on such $n$-dimensional feature vectors ($n$ is the number of checkpoints used) to predict the correctness of the final prediction. We use the predicted probability from this classifier as the calibrated confidence value. These variants are compared in Table~\ref{tab:consistency}. Interestingly, the binary implementation of \name{} turns out to be the most important than all other baselines and variants. 

% \sewon{I propose deleting the classifier-based. It is not super intuitive, and seems like it does not outperform the binary approach.} \chen{I suggest even removing frequency based approach, it takes some time to understand, and does not help \metric{}, maybe put these variants along with number of checkpoints in appendix}. \chenglei{I don't mind, my only concern is whether reviewers will be like why don't you try more complicated ways than binary.}

\begin{table*}[t]
\small
\setlength\tabcolsep{5pt}
\centering
\begin{tabular}{ c c c c c c | c c c c c} 
& \multicolumn{5}{c}{IID (NQ)} & \multicolumn{5}{c}{OOD (HotpotQA)} \\
 \toprule
 Calibrator &  
 EM & ECE\textsubscript{interval} & ICE\textsubscript{pos} & ICE\textsubscript{neg} & MacroCE & EM & ECE\textsubscript{interval} & ICE\textsubscript{pos} & ICE\textsubscript{neg} & MacroCE
 \\
 \midrule
 \multicolumn{11}{c}{\textbf{\em Non-Consistency Baselines}} \\
  \midrule
  \multicolumn{8}{l}{\textbf{\em No Calibration}} \\
   Joint & 35.24 & 30.41 & 27.17 & 61.74 & 44.46 & 24.50 & 44.40 & 25.38 & 67.04 & 46.21 \\
    Pipeline & 36.34 & 50.08 & 6.93 & 82.63 & 44.78  & 22.78 & 61.22 & 12.48 & 82.95 & 47.72 \\
    \midrule
\multicolumn{8}{l}{\textbf{\em Binary Baseline}} \\
    Joint & 35.24 & 38.03 & 53.22 & 29.77 & 41.50 & 24.50 & 38.62 & 55.81 & 33.05 & 44.43 \\
    Pipeline & 36.34 & 33.68 & 46.34 & 26.46 & 36.40  & 22.78 & 38.82 & 55.45 & 33.92 & 44.68 \\
\midrule
\multicolumn{8}{l}{\textbf{\em Average Baseline}} \\
    Joint & 35.24 & 2.00 & 64.25 & 35.75 & 50.00 & 24.50 & 11.25 & 64.25 & 35.75 & 50.00 \\
    Pipeline & 36.34 & 0.03 & 63.66 & 36.34 & 50.00 & 22.78 & 13.56 & 63.66 & 36.34 & 50.00 \\
\midrule
 \multicolumn{11}{c}{\textbf{\em Consistency Calibration with n=17}} \\
\midrule
\multicolumn{8}{l}{\textbf{\em Consistency Calibration (Binary)}} \\
Joint &  35.24 & 31.91 & 26.10 & 35.07 & 30.59 & 24.50 & 40.37 & 34.29 & 42.35 & 38.32 \\
Pipeline & 36.34 & 31.30 & 25.91 & 34.38 & 30.15 & 22.78 & 43.25 & 37.00 & 45.09 & 41.05 \\
\midrule
\multicolumn{8}{l}{\textbf{\em Consistency Calibration (Frequency)}} \\
Joint & 35.24 & 44.46 & 26.39 & 44.68 & 35.53 & 24.50 & 46.21 & 31.72 & 49.89 & 40.81 \\
Pipeline & 36.34 & 44.78 & 26.04 & 43.78 & 34.91 & 22.78 & 47.72 & 32.34 & 51.87 & 42.11 \\
\midrule
\multicolumn{8}{l}{\textbf{\em Consistency Calibration (Classifier)}} \\
Joint & 35.24 & 31.63 & 27.28 & 34.00 & 30.64 & 24.50 & 40.34 & 35.56 & 41.90 & 38.73 \\
Pipeline & 36.34 & 33.16 & 25.84 & 37.34 & 31.59 & 22.78 & 44.60 & 31.81 & 48.38 & 40.09 \\
\midrule
 \multicolumn{11}{c}{\textbf{\em Consistency Calibration with n=9}} \\
\midrule
\multicolumn{8}{l}{\textbf{\em Consistency Calibration (Binary)}} \\
Joint & 35.24 & 34.07 & 24.29 & 39.39 & 31.84 & 24.50 & 42.88 & 30.61 & 46.86 & 38.74 \\
Pipeline & 36.34 & 32.52 & 23.78 & 37.51 & 30.65 & 22.78 & 45.36 & 32.30 & 49.22 & 40.76 \\
\midrule
\multicolumn{8}{l}{\textbf{\em Consistency Calibration (Frequency)}} \\
Joint & 35.24 & 23.17 & 25.06 & 46.05 & 35.56 & 24.50 & 33.91 & 30.34 & 51.29 & 40.81  \\
Pipeline & 36.34 & 21.84 & 24.94 & 44.71 & 34.83 & 22.78 & 35.93 & 30.80 & 52.98 & 41.89 \\
\midrule
\multicolumn{8}{l}{\textbf{\em Consistency Calibration (Classifier)}} \\
Joint  & 35.24 & 32.19 & 27.67 & 34.64 & 31.16  & 24.50 & 41.11 & 36.71 & 42.53 & 39.62 \\
Pipeline  & 36.34 & 31.36 & 26.37 & 34.20 & 30.29 & 22.78 & 43.70 & 36.01 & 45.97 & 40.99 \\
\midrule
 \multicolumn{11}{c}{\textbf{\em Consistency Calibration with n=5}} \\
\midrule
\multicolumn{8}{l}{\textbf{\em Consistency Calibration (Binary)}} \\
Joint & 35.24 & 33.07 & 26.02 & 37.43 & 31.72 & 24.50 & 41.33 & 34.41 & 43.58 & 38.99 \\
Pipeline & 36.34 & 32.49 & 26.30 & 36.03 & 31.16 & 22.78 & 43.67 & 36.51 & 47.02 & 41.77 \\
\midrule
\multicolumn{8}{l}{\textbf{\em Consistency Calibration (Frequency)}} \\
Joint & 35.24 & 26.70 & 22.74 & 48.21 & 35.48 & 24.50 & 35.86 & 28.22 & 53.07 & 40.65 \\
Pipeline & 36.34 & 25.44 & 22.54 & 46.92 & 34.73  & 22.78 & 39.05 & 29.39 & 55.08 & 42.24 \\
\midrule
\multicolumn{8}{l}{\textbf{\em Consistency Calibration (Classifier)}} \\
Joint & 35.24 & 33.41 & 26.02 & 37.43 & 31.72 & 24.50 & 41.33 & 34.41 & 43.58 & 38.99 \\
Pipeline & 36.34 & 31.16 & 31.78 & 30.81 & 31.30 & 22.78 & 40.43 & 43.31 & 39.58 & 41.45 \\
 \bottomrule
\end{tabular}
\caption{Results of using consistency calibration with a larger number of checkpoints ($n=\{9,17\}$). The difference with using $n=5$ is small. Note that $n$ represents the total number of checkpoints apart being used. We find that the impact of using different $n$ is small. We reported results of using $n=5$ in the main paper.}
\label{tab:consistency_8}
\end{table*}

\begin{figure}[t]
 \centering
 \includegraphics[trim=0.6cm 0.5cm 1.0cm 0.5cm,clip=true,width=1.01\linewidth]{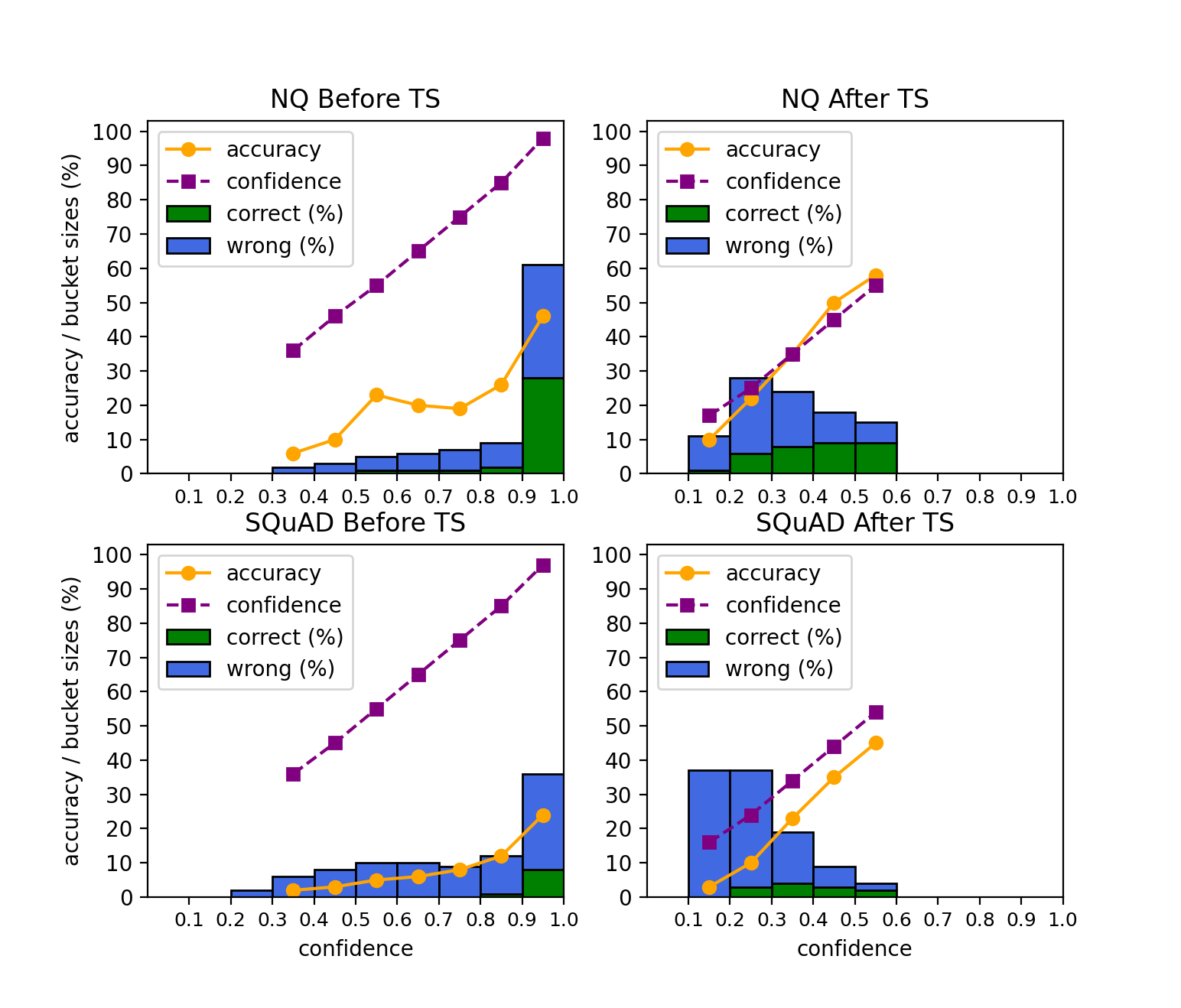}
\caption{Bucketing distribution of predictions on \abr{nq} and \abr{HotpotQA}.
The x-axis represents the confidence range of each bucket, the y-axis represents the average answer accuracy for the dashed line plot and represents the relative bucket sizes for the histogram.
Similar flaws hold true for these two datasets: after temperature scaling, all predictions' confidence values are scaled to become closer to the overall answer accuracy,
and correct (green bars) and wrong predictions (blue bars) are mixed in the same buckets, making it hard to distinguish.}
\label{fig:more_ece_illustration}
 \end{figure}

 \begin{figure}[t]
 \centering
 \includegraphics[width=1.\linewidth]{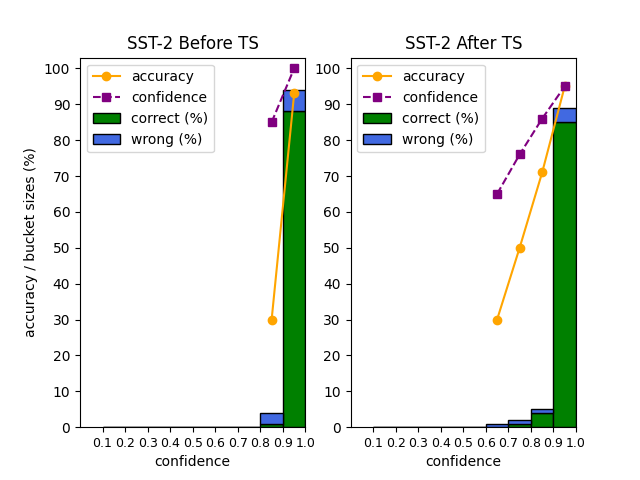}
\caption{Bucketing distribution of predictions on SST-2~\cite{Socher2013RecursiveDM}, a binary sentiment analysis dataset.
We can see the same trend as on \abr{nq} and \abr{HotpotQA} - all predictions' confidence are very close. In fact, on SST-2, all predictions have high confidence both before and after temperature scaling, making it hard for users to identify wrong predictions.}
\label{fig:more_ece_illustration_sst2}
 \end{figure}

\section{Impact of Checkpoint Numbers in \name{}}\label{app:impact-checkpoint-number}

In the main paper we saved a total number of $n=5$  checkpoints during training for \name{}. In order to understand the impact of this hyper-parameter $n$, we experiment with $n=\{9, 17\}$ and report the results in Table~\ref{tab:consistency_8}. We observe that the impact of different $n$ is very small.

\section{Illustration of \abr{ece} Flaws on More Datasets}
\label{app:more_ece}

In the main paper we illustrated the flaws of \abr{ece} with a case study on \abr{HotpotQA}. Here we additional present visualizations of calibration results on \abr{nq} (in-domain) and \abr{SQuAD} (\abr{ood}). As shown in Figure~\ref{fig:more_ece_illustration}, the flaws of \abr{ece} as described in the main paper hold true for these datasets as well, validating the generality of our conclusions. 

In addition to \abr{qa}, we also present results on a sentiment analysis dataset in Figure~\ref{fig:more_ece_illustration_sst2}. We observe the same trends that all predictions have similar confidence scores, making it difficult to identify the wrong predictions.

\clearpage

\end{document}